\documentclass[lettersize,journal]{IEEEtran}
\usepackage{amsmath,amsfonts}
\usepackage{algorithm}
\usepackage{algorithmicx}
\usepackage{algpseudocode}
\usepackage{array}
\usepackage[caption=false,font=normalsize,labelfont=sf,textfont=sf]{subfig}
\usepackage{textcomp}
\usepackage{stfloats}
\usepackage{url}
\usepackage{verbatim}
\usepackage{graphicx}
\usepackage{multirow,booktabs}
\usepackage{xcolor}
\usepackage{threeparttable}
\usepackage{bm}
\usepackage{hyperref}
\def\BibTeX{{\rm B\kern-.05em{\sc i\kern-.025em b}\kern-.08em
    T\kern-.1667em\lower.7ex\hbox{E}\kern-.125emX}}
\usepackage{balance}

\usepackage{setspace} 

\usepackage[numbers,sort&compress]{natbib}

\makeatletter
\def\thanks#1{\protected@xdef\@thanks{\@thanks
        \protect\footnotetext{#1}}}
\makeatother

\begin{document}
\title{LSGDDN-LCD: An Appearance-based Loop Closure Detection using Local Superpixel Grid Descriptors and Incremental Dynamic Nodes}


\author{Baosheng Zhang, Yanpeng Dong, Xianyu Qi, Mengfan Li, and Xiaoguang Ma}

\thanks{Baosheng Zhang, Yanpeng Dong, Xianyu Qi, Mengfan Li are with the Beijing Institute of Mechanical Equipment, Beijing 100854, China (e-mail: zhangbaosheng0503@163.com; dongyanpengcn@sina.com; qixianyu@buaa.edu.cn; bitlmf@126.com).}

\thanks{Xiaoguang Ma is with the College of Information Science and Engineering, Northeastern University, 110819 Shenyang, China (Xiaoguang Ma is the corresponding author, e-mail: maxg@mail.neu.edu.cn).}

\maketitle
\begin{abstract}
    Loop Closure Detection (LCD) is an essential component of visual simultaneous localization and mapping (SLAM) systems. It enables the recognition of previously visited scenes to eliminate pose and map estimate drifts arising from long-term exploration. However, current appearance-based LCD methods face significant challenges, including high computational costs, viewpoint variance, and dynamic objects in scenes. 
This letter introduced an online appearance based LCD using local superpixel grids descriptor (LSGD) and dynamic nodes, i.e, LSGDDN-LCD, to find similarities between scenes via hand-crafted features extracted from the LSGD. Unlike traditional Bag-of-Words (BoW) based LCD, which requires pre-training, we proposed an adaptive mechanism to group similar images called $\textbf{\textit{dynamic}}$ $\textbf{\textit{nodes}}$, which incrementally adjusted the database in an online manner, allowing for efficient and online retrieval of previously viewed images without need of the pre-training.
Experimental results confirmed that the LSGDDN-LCD significantly improved LCD precision-recall and efficiency, and outperformed several state-of-the-art (SOTA) approaches on multiple typical datasets, indicating its great potential as a generic LCD framework.
\end{abstract}


\section{Introduction}
\IEEEPARstart{T}{he} development of autonomous robotics and augmented reality has garnered significant attention on simultaneous localization and mapping (SLAM), which utilizes various sensors to construct precise maps of unknown environments and incrementally localizes itself without any prior information. 
Practically, visual SLAM (vSLAM) systems are particularly affected by viewpoint variance, dynamic objects, and layout changes in real scenes . These factors can cause accumulated errors during long-term and large-scale mapping and localization processes. To cope with these challenges, loop closure detection (LCD) module has been introduced into vSLAM systems to correctly identify previously-seen scenes, reduce drift, and improve the accuracy of the vSLAM systems.   

Due to its cost-effectiveness and broad applicability, appearance-based methods have emerged as a popular technique for detecting loop closures. The LCD methods usually consist of two stages: feature extraction and candidate frame selection, wherein the feature extraction aims to represent scenes effectively and robustly and can be realized by using binary descriptors such as BRIEF \cite{calonder2010brief}, ORB \cite{Bescos2018DynaSLAM}, LDB \cite{yang2013local}, and AKAZE \cite{alcantarilla2011fast}, as well as traditional point feature detectors like SURF \cite{bay2006surf} and SIFT \cite{lowe2004distinctive}. Visual words are then generated via quantization of the space of local descriptors. For example Bag-of-Words (BoW) \cite{nister2006scalable} approach utilizes widely-used term frequency-inverse document frequency (TF-IDF) technique to construct a visual word histogram and combines it with inverted index method to rapidly calculate similarity between current query images and previous images for identifying potential loop closure pairs, i.e., candidate frames.

Despite the effectiveness of existing BoW-based LCD methods, several challenges still need to be addressed. Firstly, conventional BoW framework relies solely on the distribution of local hand-crafted features and neglects the spatial relationships between features in images, making it vulnerable to severe scene changes. Secondly, the contribution of a pre-trained module for database of visual vocabulary can be time-consuming, leading to false loop detection when querying datasets with different appearance types from the training image sets. These issues call for further investigation efforts to enhance the performance and robustness of the LCD for real application. Specially, it is important to investigate image features that consider geometric structures and local characteristics, along with efficient online image retrieval methods.

In order to address the issues mentioned above, this letter presented an appearance-based LCD using local superpixel grid descriptors and incremental dynamic nodes, LSGDDN-LCD, an online and efficient visual LCD approach, suitable for handling severe scene and viewpoint changes. The proposed approach employed Superpixel Grid (SG) algorithm to segment scenes into multiple semantic regions, each containing local information, wherein SG features were extracted based on pixel intensity to encode each image and enhance system's robustness and performance. To further improve efficiency, we introduced $dynamic$ $nodes$ to group images with similar appearance and reduce processing time for the LCD. Our experiments demonstrated that this approach achieved significant improvements in both precision-recall performance and efficiency, especially when dealing with large-scale datasets.

To summarize, the main contributions of this letter are as follows:
\begin{itemize}
    \item We proposed a novel patch level feature extraction approach, based on SG segmentation and local spatial characteristics to represent images, by mitigating the adverse effects of significant appearance changes to improve both precision and recall performance.
    \item A novel concept, $dynamic$ $nodes$ was proposed for dynamically grouping similar visited images and selected candidate frames. This approach did not require offline or online training stages nor the use of a trained large-scale visual dictionary, reducing computing time significantly.
    \item A new NEU indoor dataset was recorded, including strong lighting changes and severe appearance changes, with typical loop-closure. It is available at \url{https://drive.google.com/drive/folders/1-tRfQ3cKriTVYb2mmEyF45Ij1mHpRjkX?usp=sharing}
    \item We introduced a complete appearance-based LCD approach, using local superpixel grid descriptors and incremental dynamic nodes (LSGDDN-LCD), which was found to be able to improve precision-recall performance and efficiency over other SOTA LCD approaches in most scenes.
\end{itemize}
The rest of this letter was organized as follows. Section \ref{sec:2} provided a brief overview of related works of the LCD. Section \ref{sec:3} described proposed algorithm in details. Comparative experiments and analysis were presented in Section \ref{sec:4}. Finally, we summarized our findings and discussed future work in Section \ref{sec:5}.

\section{Related Work}
\label{sec:2}
Appropriate feature selection and rational utilization of extracted features are critical for achieving accurate identification of visited location scenes, as well as robustness against strong viewpoint variations and appearance changes in LCD methods. This section provided a concise overview of algorithms related to image feature extractors and other relevant approaches utilized in the LSGDDN-LCD.

\subsection{Hand-Crafted Feature Extractors}
Image representation is a crucial aspect of appearance-based LCD algorithms. 
Most methods used feature sets to reduce image dimensionality and  represent scenes efficiently. For example, fast appearance-based mapping (FAB-MAP) \cite{cummins2008fab} extracted SURF point features from query images and clustered them into visual words using k-means in low-dimensional space. A Chow-Liu tree approximated the probabilities of visual word co-occurrence for fast similarity evaluation between query and database images. Other feature extractors, such as BRIEF \cite{calonder2010brief}, ORB \cite{rublee2011orb}, SITF \cite{lowe2004distinctive}, and KAZE \cite{alcantarilla2012kaze} were also proposed for various scenes. However, these hand-crafted feature extractors had several limitations, including high computational cost, low robustness, and susceptibility to matching errors caused by noise and deformation in images. 

\subsection{Feature Integration and Encoding Optimization}
To enhance feature performance, researchers focused on integrating extracted features to preserve image characteristics and enable fast and accurate LCD. Although the BoW models grouped similar visual words for direct matching of BoW descriptors, visual words generated by clustering image features using k-means or k-medoids do not consider geometric information between features in images \cite{angeli2008fast, galvez2012bags, nicosevici2012automatic, khan2015ibuild}, limiting their expressive ability in complex scenes. 

To cope with this, researchers started to introduce temporal or spatial constraints into the LCD, and attempted to improve querying performance by using vectorized representations such as vectors of local aggregated descriptors (VLAD) \cite{arandjelovic2016netvlad, torii201524} and fisher vectors (FV) \cite{sanchez2013image}, or by employing voting approaches such as hierarchical navigable small world (HNSW) \cite{an2019fast}. Although these works aimed to encode and match visual features more effectively to enhance LCD performance, they were based on unstable point features and required a training stage, and a different approach by extracting patch-level features and building the dictionary in an online manner could be promising to address this problem.

\subsection{Addressing Appearance Changes in Place Recognition}
Although research on place recognition in changing environments  developed rapidly, no consistent solution was available yet to handle practical challenges of detecting severe appearance and viewpoint changes. Some researchers tried to use fixed patches to detect repetitive features. Naseer $et$ $al.$ \cite{naseer2014robust} proposed a graph approach that formulated image matching as a minimum cost flow problem in a data association graph. To calculate image similarity, they used dense and regular grids of histogram of oriented gradients (HOG) descriptors and generated multiple route matching hypotheses by optimizing network flows. Milford $et$ $al.$ \cite{milford2013towards} selected candidate frames based on whole image matching and verified correct loop-closure using local image regions. Although this method could reduce false positive matches, it was not able to increase match numbers when there were viewpoint changes, leading to a large number of missed matches in the process of loop detection. 


\section{Technical Approach}
\begin{figure}[!t]  
    \centering
    \includegraphics[width=3.4in]{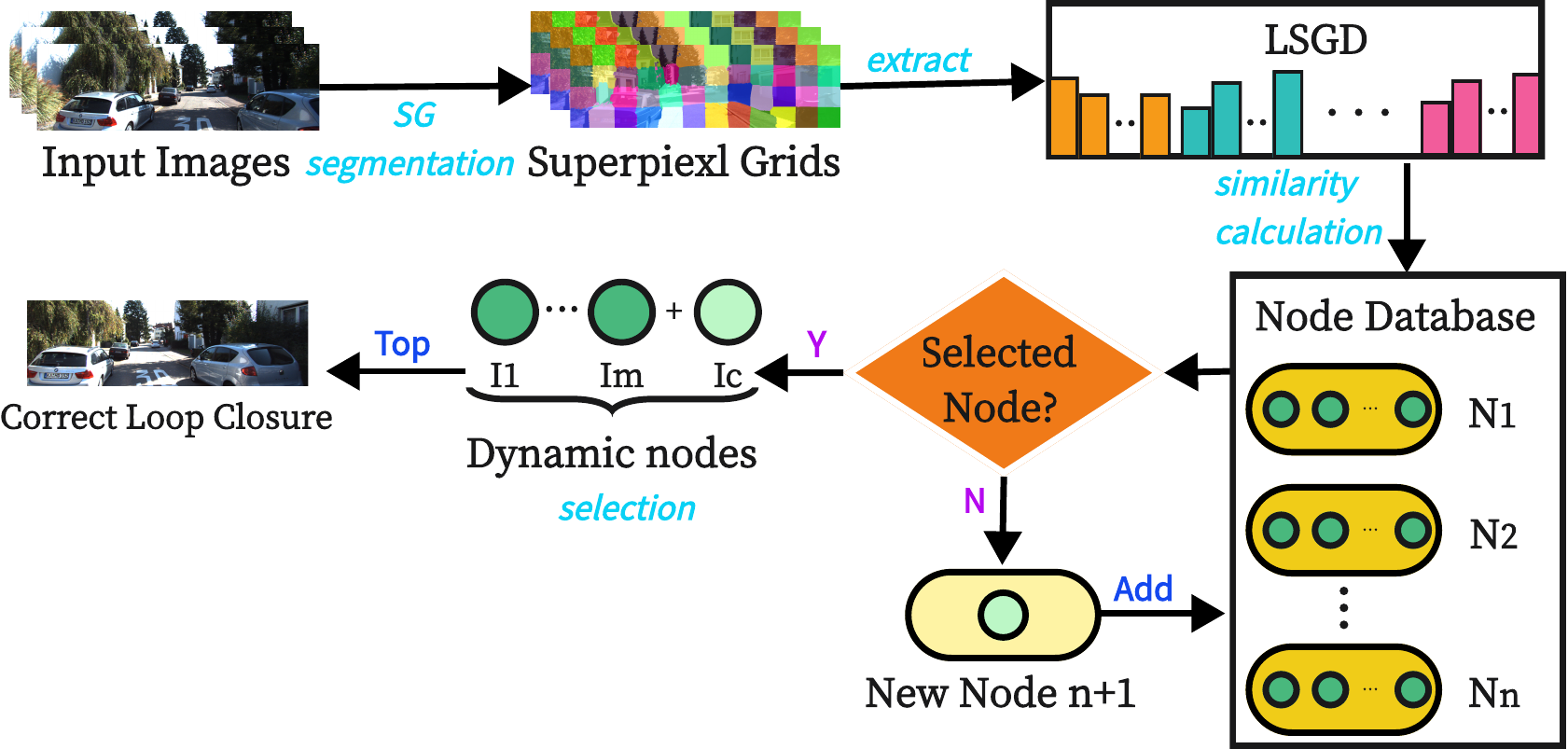}
    \caption{Framework of the LSGDDN-LCD.} 
    \label{fig:ov}
\end{figure} 
\label{sec:3}
\subsection{Overview}
Figure \ref{fig:ov} provided a framework of the LSGDDN-LCD approach, which comprised four components, i.e., superpixel grid (SG) segmentation based on local appearance characteristics, extracting LSGD features from the SGs, similarity calculation by the LSGD, and dynamic node selection based on the similarity between images. 

Our approach involved several key components. Firstly, we implemented the SG segmentation, which divided current scenes into M $\times$ N irregular grids with local scene information, i.e., SG. This method was inspired by the linear iterative clustering (SLIC) algorithm \cite{achanta2012slic} and helped to address the adverse effects of changes in the environment on implementation effectiveness.
In Section \ref{sec:32}, we would provide details of local feature extraction based on the SGs, involving extracting features from both the segmented SGs as mentioned above and the pixel intensities of the images. Furthermore, we illustrated the advantages of SG features over traditional point features from various perspectives, and the similarity between query and database images was also calculated. To efficiently search previously visited scenes and select candidate frames, we introduced $dynamic$ $nodes$ in Section \ref{sec:33}. By identifying the image with the highest similarity to the query image in the selected candidate frames, we could determine the correct loop-closure.
\subsection{Superpixel Grid (SG) Segmentation}
\label{sec:31}
\begin{figure*}[!t]  
    \captionsetup[subfloat]{labelsep=none,format=plain,labelformat=empty}
    \centering    
    \subfloat 
    {
        \includegraphics[width=0.22\textwidth]{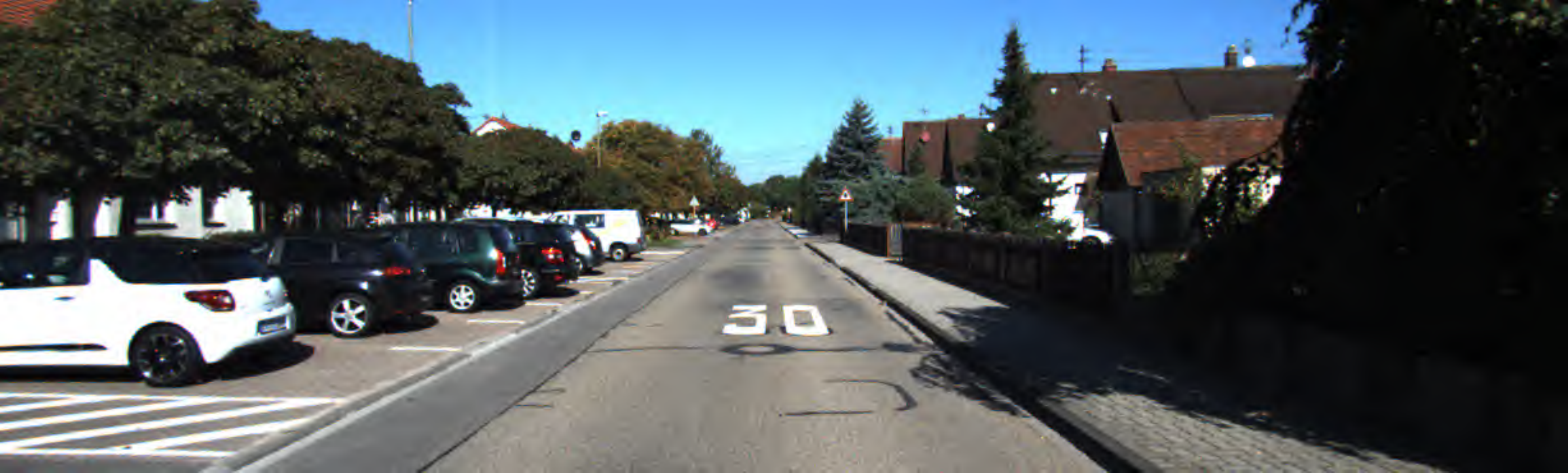}
    }
    \subfloat 
    {
        \includegraphics[width=0.22\textwidth]{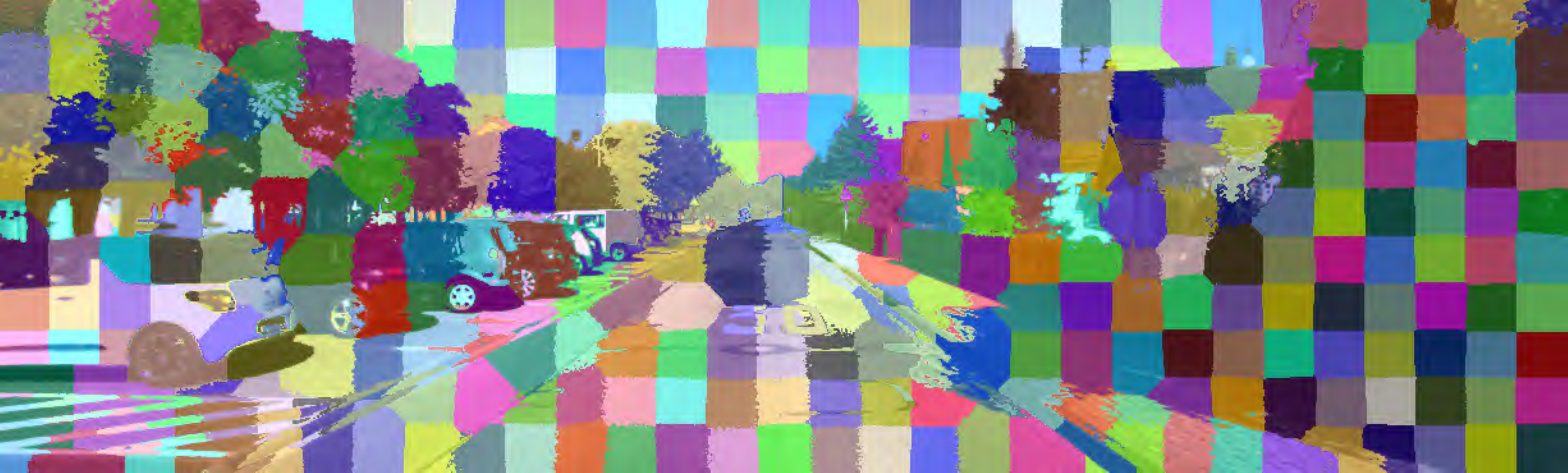}
    }
    \subfloat 
    {
        \includegraphics[width=0.22\textwidth]{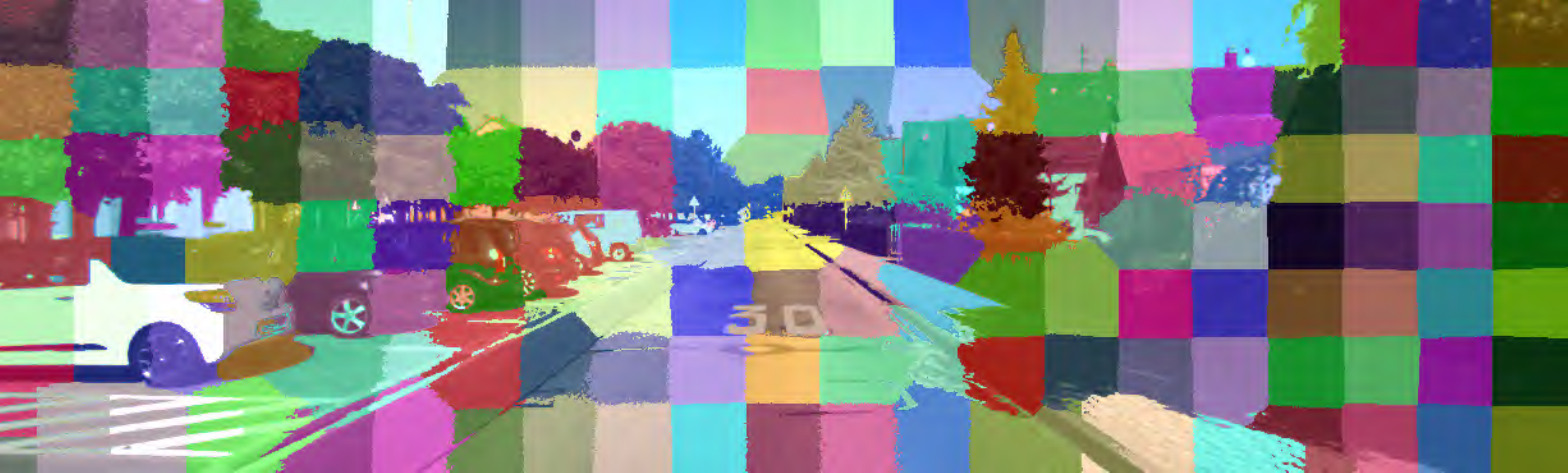}
    }
    \subfloat 
    {
        \includegraphics[width=0.22\textwidth]{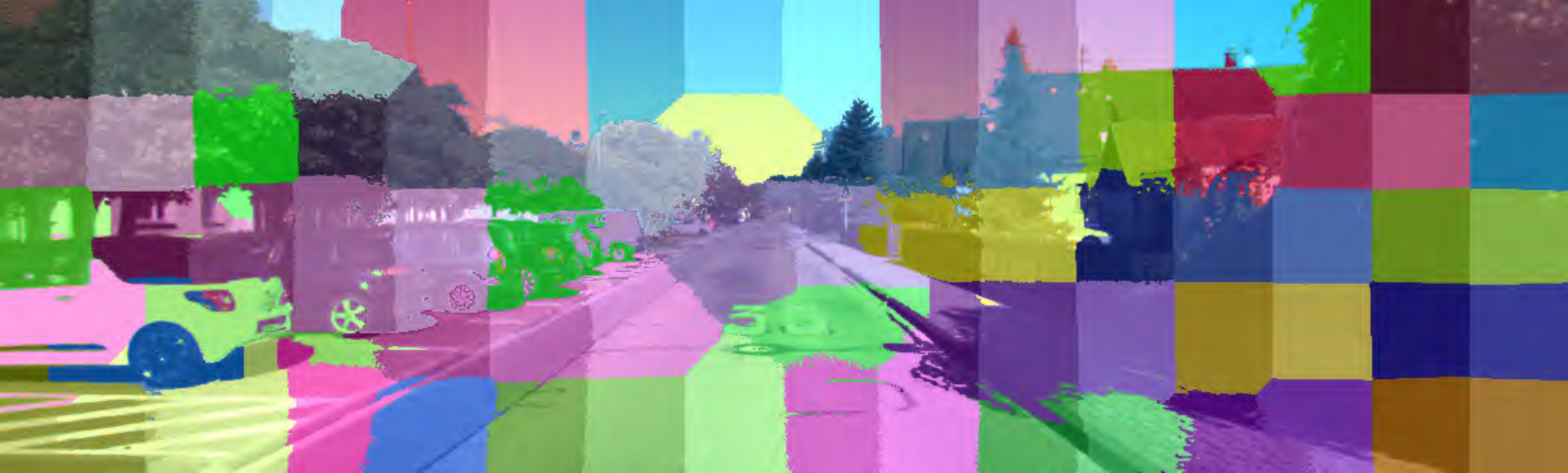}
    }
   
    \subfloat[(a) Example images] 
    {
        \includegraphics[width=0.22\textwidth]{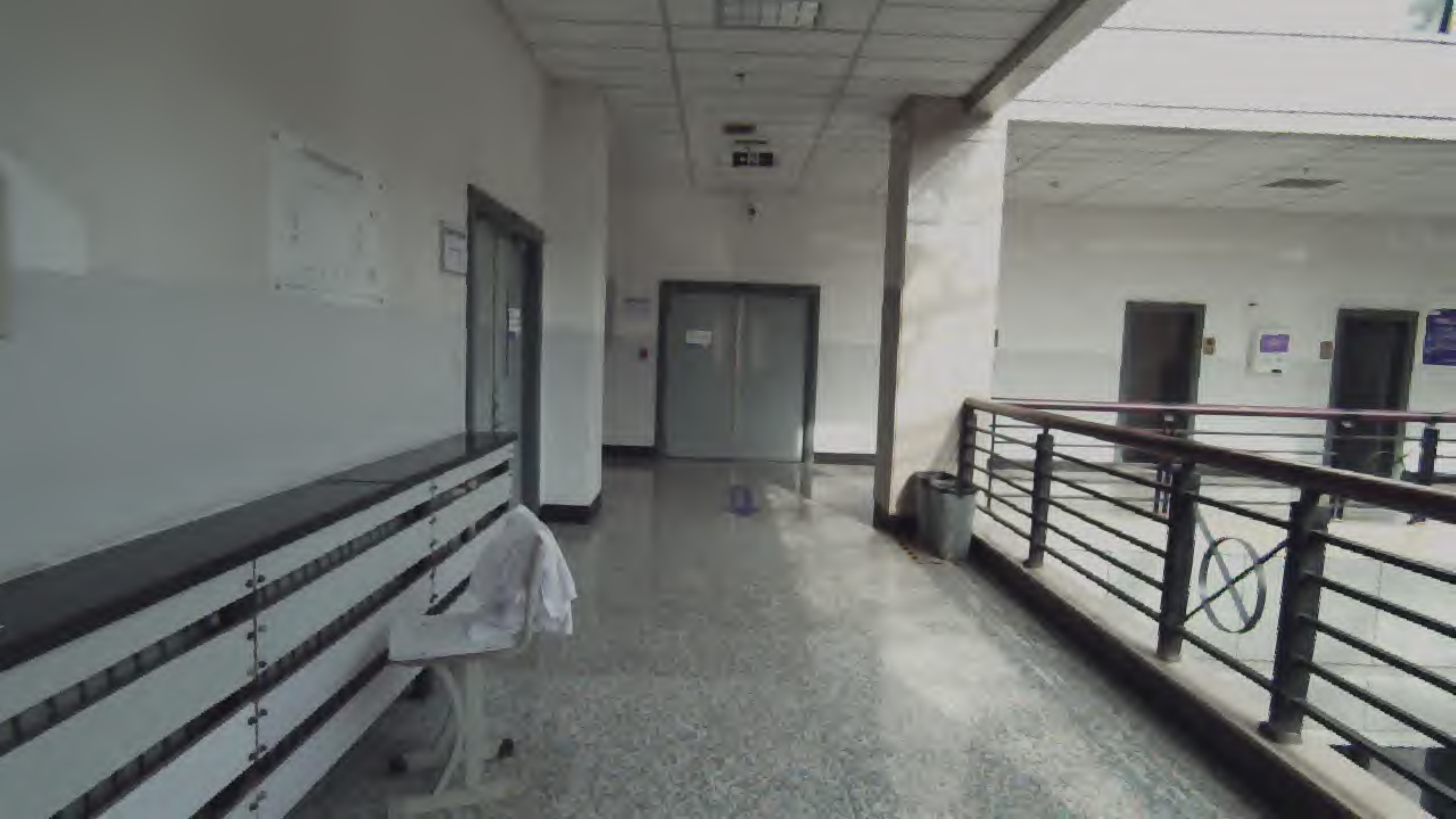}
    }
    \subfloat[(b) $S_p$ = 20] 
    {
        \includegraphics[width=0.22\textwidth]{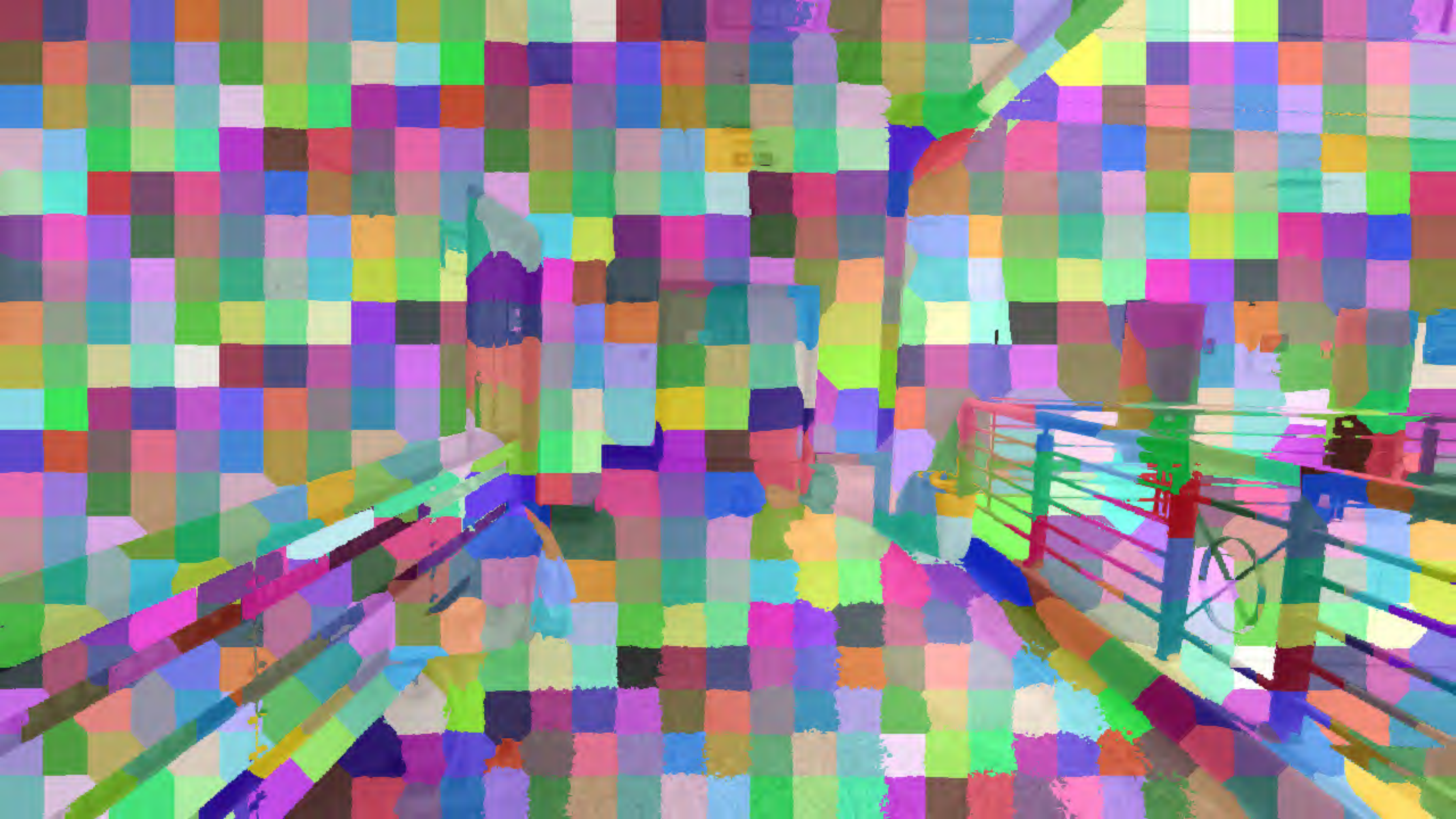}
    }
    \subfloat[(c) $S_p$ = 30] 
    {
        \includegraphics[width=0.22\textwidth]{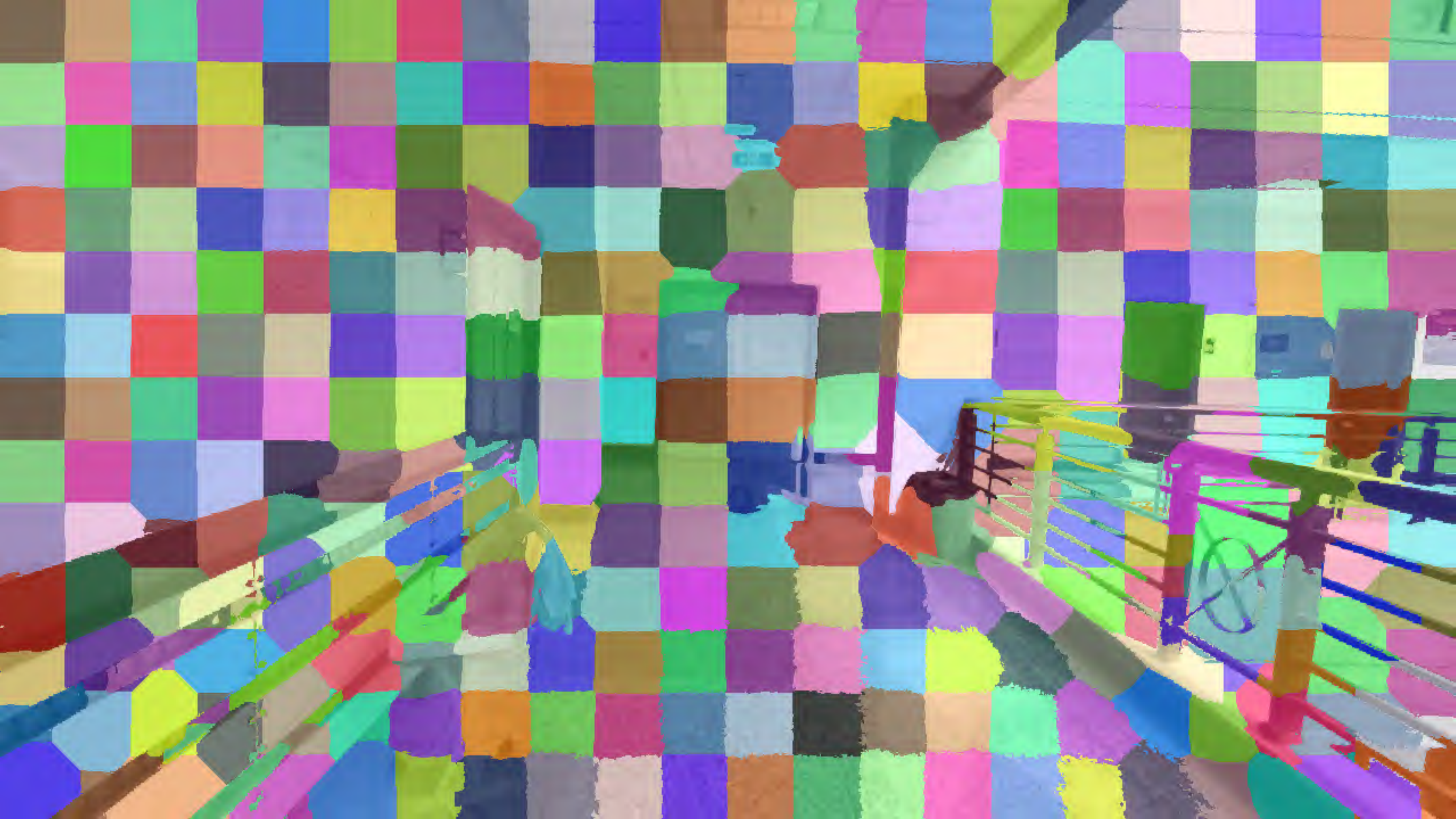}
    }
    \subfloat[(d) $S_p$ = 40] 
    {
        \includegraphics[width=0.22\textwidth]{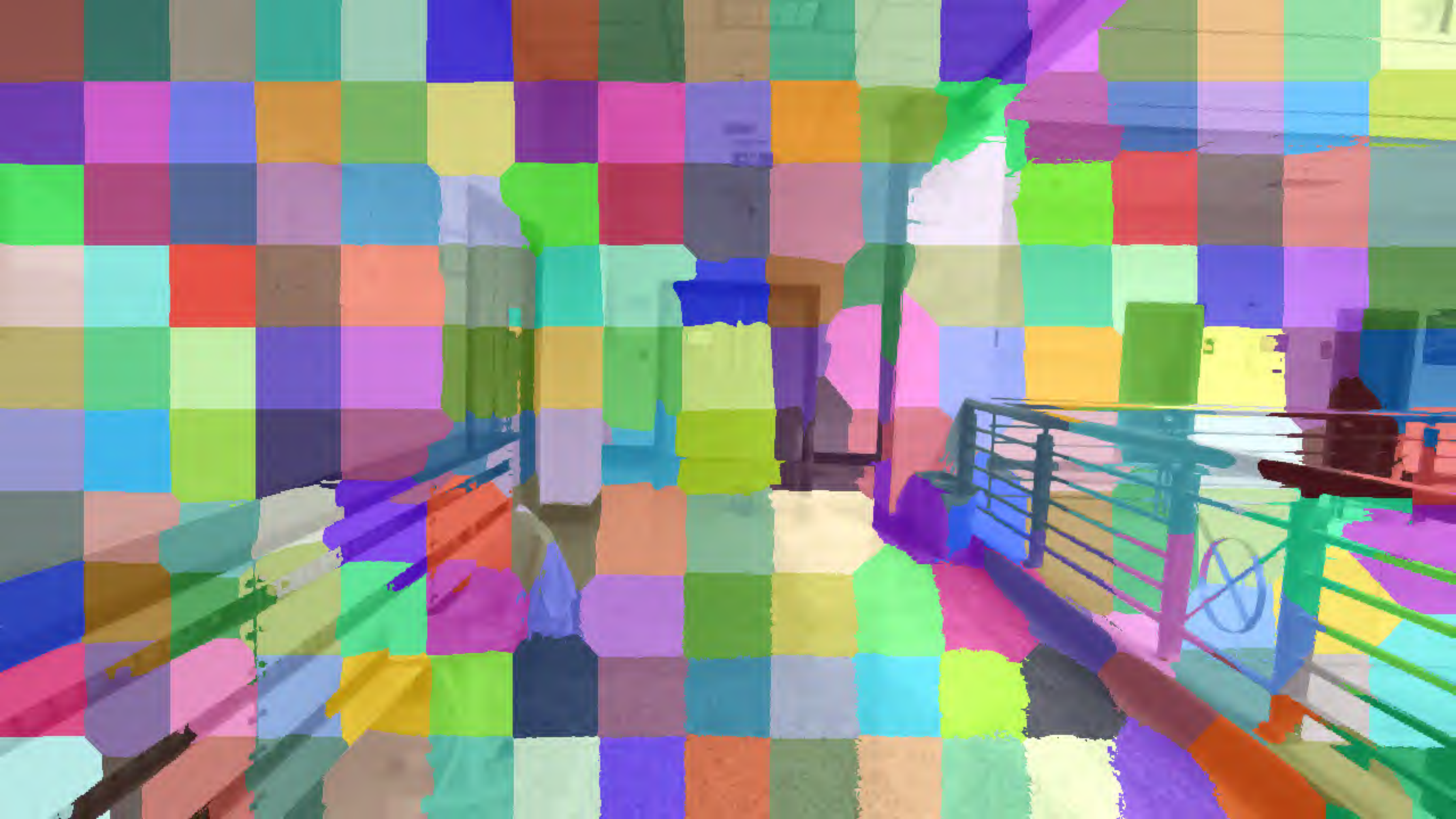}
    }
    \caption{The upper and lower lines of the images, depicted outdoor scenes (from the KITTI dataset) and indoor scenes (from the NEU dataset), respectively. Superpixel Grid Segmentation was performed on these images using various segmentation scales, i.e., $S_p$ . The resulting segmentation highlighted the variability in optimal $S_p$ selection for different scene categories.} 
    \label{fig:tec1}  
\end{figure*}
When a scene contains dynamic objects or severe viewpoint changes, the performance of an LCD system may significantly degrade. In feature-matching methods, local features had been demonstrated to be more robust to appearance changes over global features and are therefore widely used in image representation \cite{garcia2018ibow} \cite{park2019robust}. In this letter, we proposed a SG segmentation method based on the SLIC \cite{achanta2012slic} framework, wherein images were divided into irregular grids based on the local features to convert scene information into a more robust form.

To obtain SGs with high resilience, we first converted images to grayscale and initialized them with fixed grids of various sizes, wherein uniformly distributed center points of the fixed grids were selected as the initial centers of the SGs before performing SG segmentation. In this process, the segmentation scale, which refered to the size of the initialized fixed grids, was defined as $S_p$. Additionally, we defined the fuse distance, denoted as $F_d$, between the center and other points in the SG. It could be calculated by 
\begin{equation*}
    \begin{aligned}
        & d_e =\sqrt{(x_c-x_i)^{2} + (y_c-y_i)^{2} }\\
        & d_i =\sqrt{(p_c-p_i)^{2} } \\
        & F_d =\sqrt{\left(\frac{d_e}{n}\right )^{2} + \left (\frac{d_i}{m}\right )^{2}}  
    \end{aligned}
\end{equation*}
where $d_e$ and $d_i$ represented the Euclidean distance and pixel intensity difference between the SG center and other points in the same SG. The variables $(x_c, \, y_c)$ and $(x_i, \, y_i)$ denoted the locations of the center and local region points in the SG, and $p_c$ and $p_i$ represented their respective pixel intensities. The range of values for $x_i$ and $y_i$ were $(x_c-S_p,\, x_c+S_p)$ and $(y_c-S_p, \, y_c+S_p)$, respectively, and the center point of each grid was iteratively updated using the k-means algorithm until robust SGs were obtained.

Scene segmentation can be performed at multiple scales to address various types of scenes. Generally, more precise segmentation can better preserve image information. However, it also required more running time and memory resources. Figure \ref{fig:tec1} showed an example image with different values of $S_p$ for both outdoor and indoor scenes. It was obvious that outdoor open scenes, such as those shown in the upper line, were better suited for larger values of $S_p$. Conversely, complex indoor scenes, like those displayed in the lower line, worked better with a smaller value of $S_p$.
\subsection{Local superpixel grid descriptor (LSGD)}
\label{sec:32}
\begin{figure}[!t]  
    \centering    
    \subfloat[A current image] 
    {
        \begin{minipage}[!t]{0.45\textwidth}
            \centering          
            \includegraphics[width=\textwidth]{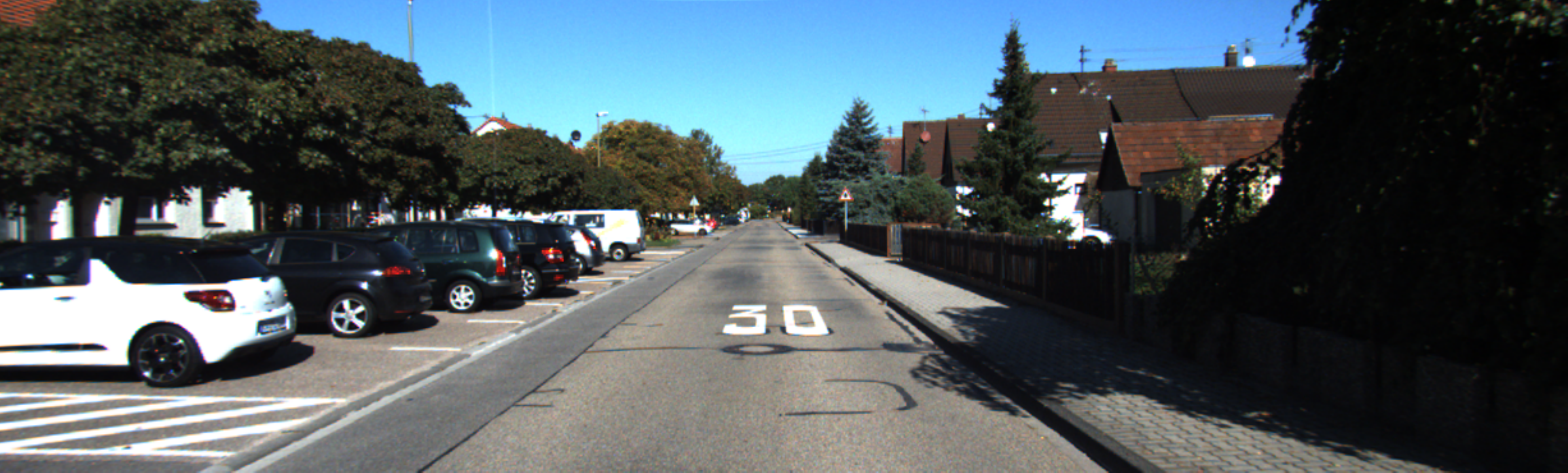}   
        \end{minipage}%
    }
    
    \subfloat[A histogram of words]
    {
        \begin{minipage}[!t]{0.45\textwidth}
            \centering          
            \includegraphics[width=\textwidth]{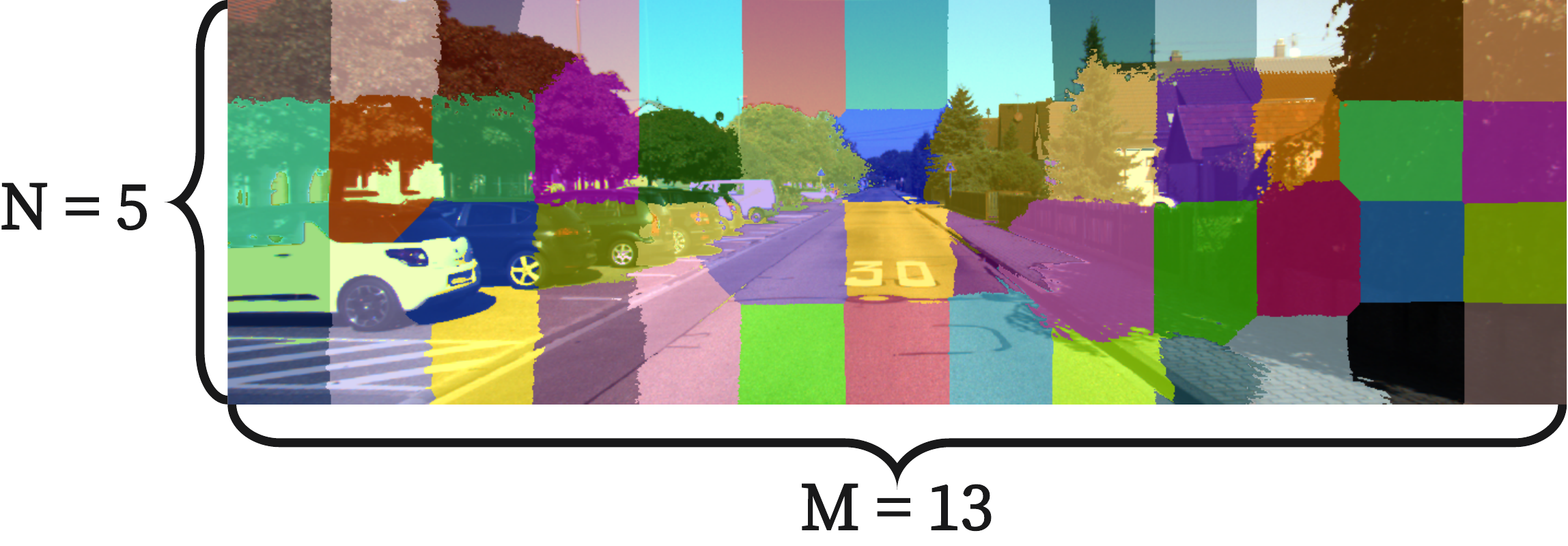}   
        \end{minipage}%
    }
    
    \subfloat[Histogram of the SG]
    {
        \begin{minipage}[!t]{0.45\textwidth}
            \centering          
            \includegraphics[width=\textwidth]{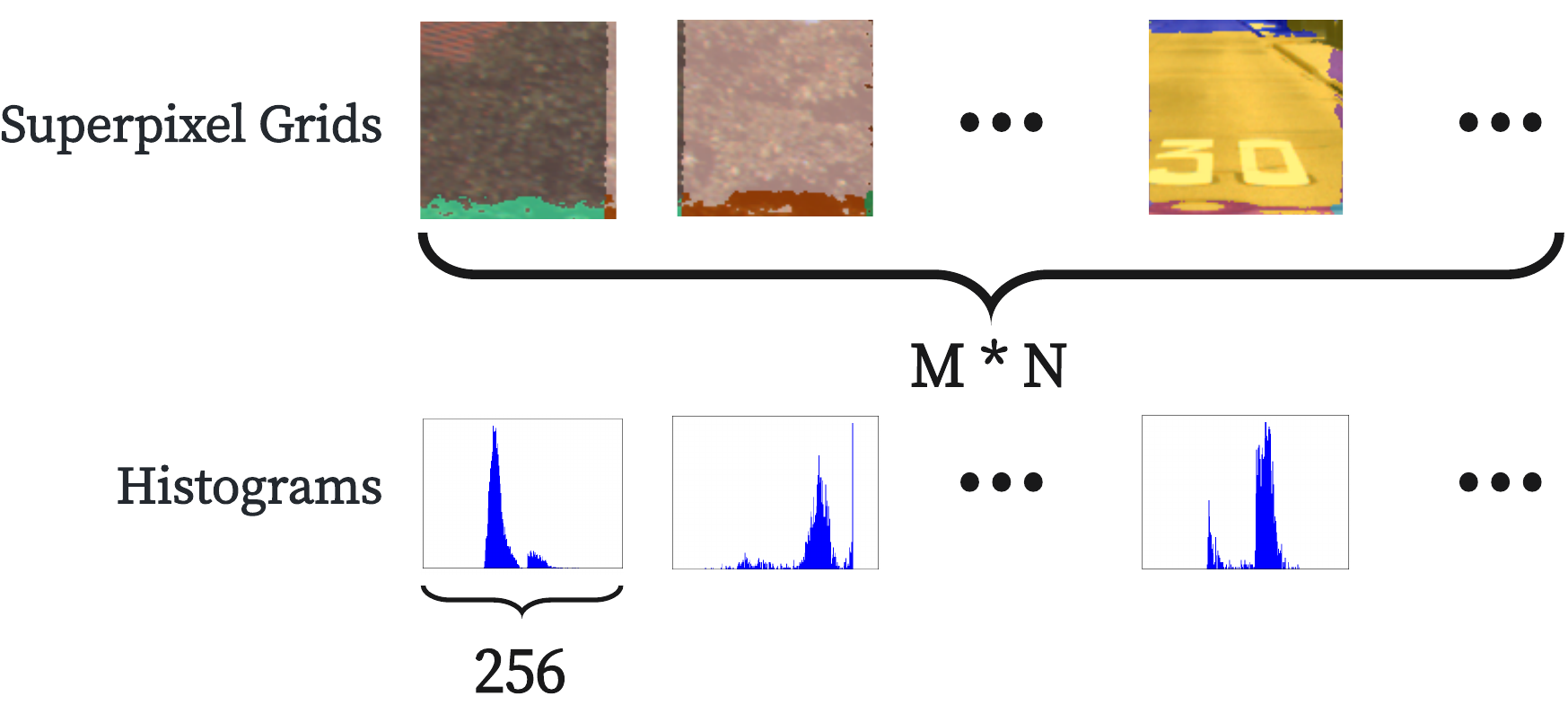}   
        \end{minipage}%
    }
    \caption{Workflow of extracting LSGD features from an input query image on the KITTI-05 dataset. The example image was shown in (a), and its SG representation obtained through SG segmentation was illustrated in (b). The LSGD descriptors were then extracted from the segmented grids, as shown in (c).} 
    \label{fig:sgf}  
\end{figure}

Repeatability and accuracy are essential in local feature extraction algorithms of the LCD, where the feature descriptors play a critical role. Moreover, the feature extractor should be time-efficient to meet real-time requirements. Therefore, we utilized SG segmentation and calculated patch-level descriptors called local superpixel grid descriptor (LSGD), which encoded the content of pixel intensity in each SGs. Specifically, the current image was represented as a histogram of words with dimensions M × N × 256, as illustrated in Figure \ref{fig:sgf}, where M and N denoted the number of grids that were segmented as described in Section \ref{sec:31}. We defined similarity score, denoted as $simScore$, to quantify the level of similarity between images. Specifically, this was achieved by computing the pixel intensity histogram of each SG and constructing a histogram with dimensions M × N × 256 to represent the image, as illustrated in Figure \ref{fig:sgf} (c), and by applying the following formula:
\begin{equation*}
    simScore = \sum_{i=1}^{M}\sum_{j=1}^{N}L_1(LSGD_{Q}(i, j), LSGD_{D}(i, j)) 
\end{equation*}
where $L_1$ represented the distance of the L1 norm, and $LSGD_Q$ and $LSGD_D$ denoted the LSGD descriptors of query and database images, respectively.

This approach effectively overcame the deficiencies of traditional feature descriptors such as SURF, SIFT, and ORB in practical application scenarios. Firstly, the LSGD efficiently encoded all pixels while preserving more information of the image without requiring complex pixel-level operations that consumed computation time. Secondly, the use of local pixel intensity based on the SGs greatly reduced the impact of dynamic objects and viewpoint changes. Furthermore, the LSGD of each SG avoided uneven distribution of features in complex scenes, thereby improving the performance of the LCD approaches. Finally, the use of patch-level descriptors reduced the occurrence of false matches, which were very common in traditional feature descriptors, greatly improving the accuracy of the LCD systems.
\subsection{Building and Selecting Dynamic Nodes}
\label{sec:33}
Most visual LCD methods based on the BoW model required a training stage. However, input scenes could be very different from those used in the training phase, leading to false loop detection. In this work, we proposed an incremental online approach to overcome this issue by creating image groups, i.e., $dynamic$ $nodes$s. Once the current frame $I_c$ met the similarity requirements of the corresponding node, the traversal process would be terminated, and the system would directly move to the current loop closure verification step. As a result, it was not necessary to traverse every image in the dataset, saving computational resources and improving efficiency of the LCD system.
\begin{algorithm}
    \caption{: Outline of the $dynamic$ $nodes$ algorithm}
    \begin{algorithmic}[1]
        \Require The LSGD descriptor of $I_c$ and all previous visited images. 
        \Ensure The selected $dynamic$ $node$.
        \For {$ i=1:n $ $ $} 
            \State $S_{i1} = SimLSGD(I_c, \, I_{i1})$
            \If{$S_{i1} > \alpha$}
                  \For {$ j=1:m $ $ $}
                    \State $S\_ave_{i}$ += $SimLSGD(I_c, \, I_{ij})$
                  \EndFor
                  \State $S\_ave_{i}$ /= $m$
                  \If{$S\_ave_{i} > \beta$}
                    \State $I_c$ insert to node $N_i$ 
                    \State \Return{$dynamic$ $node$ $N_i$}
                  \EndIf
            \EndIf
        \EndFor
        \State new $dynamic$ $node$ $N_{n+1}$ from $NodeBuilding(I_c)$
        \State \Return{$dynamic$ $node$ $N_{n+1}$}          
        \end{algorithmic}
    \label{alg:1}
\end{algorithm}

To efficiently select and construct similar image groups based on scene information, we utilized a second-order similarity estimation strategy in node selection.
For current query image $I_c$, we calculated its appearance similarity with the first image, $I_{i1}$, in all existing image groups (nodes) ranging from $N_1$ to $N_n$ based on their LSGD features, which could be extracted from the query images as described in Section \ref{sec:32}. If the similarity score $S_{k1}$ for the $k$th node was greater than the predefined threshold $\alpha$, we considered the image group as a candidate node and evaluated the similarity between $I_c$ and all images in the image group $(I_{k1}, ... I_{km})$. When the average similarity score $S\_ave_{k}$ between $I_c$ and all frames in this node reached the predefined threshold $\beta$, all images in this candidate node would join the subsequent loop-closure determination as candidate frames. Then, image $I_c$ would be added to the selected node as a new group frame to participate in analyzing and selecting similar image groups for future input of new images. Notably, if all similar image groups had been traversed and no nodes met the similarity requirement, a new node would be created, and $I_c$ would become the only frame, waiting for subsequent input images that met the requirements, to gradually extend the size of the group, as depicted in Figure \ref{fig:ov}. The detailed process of $dynamic$ $nodes$ was summarized in Algorithm \ref{alg:1}.
\subsection{Loop Closing Verification}
Following the aforementioned process, images belonging to the same node typically exhibited high similarity to each other, indicating that they were likely located in the same local spatial position. To further utilize this advantage, we could put N images with the highest similarity to the current query frame $I_c$ to participate in subsequent algorithmic processes or choose the image with the highest similarity to $I_c$ as the final loop closure frame for loop closing verification.

\section{Experimental Results and Analysis}
\label{sec:4}
To evaluate the effectiveness of the proposed LSGDDN-LCD, a series of experiments were conducted. This section analyzed the performance of the LSGD and tested the $dynamic$ $nodes$ database architecture using the LSGDDN-LCD. In addition, we presented qualitative and quantitative results to demonstrate the superior performance of the LSGDDN-LCD over other SOTA LCD algorithms.

The experimental setup included a laptop equipped with Ubuntu 18.04 operating system, an Intel i7-10510U CPU @ 1.80GHz × 8, and 8GB RAM.


\subsection{Effectiveness of the LSGD}
Before evaluating the LSGDDN-LCD, we first verified the effectiveness and efficiency of the LSGD using typical precision-recall metric and analyzed how varying segmentation scales impacted the final implementation of the system.


\label{secB1}
\begin{figure*}[!t]  
    \centering    
    \subfloat[] 
    {
        \begin{minipage}[!t]{0.31\textwidth}
            \centering          
            \includegraphics[width=\textwidth]{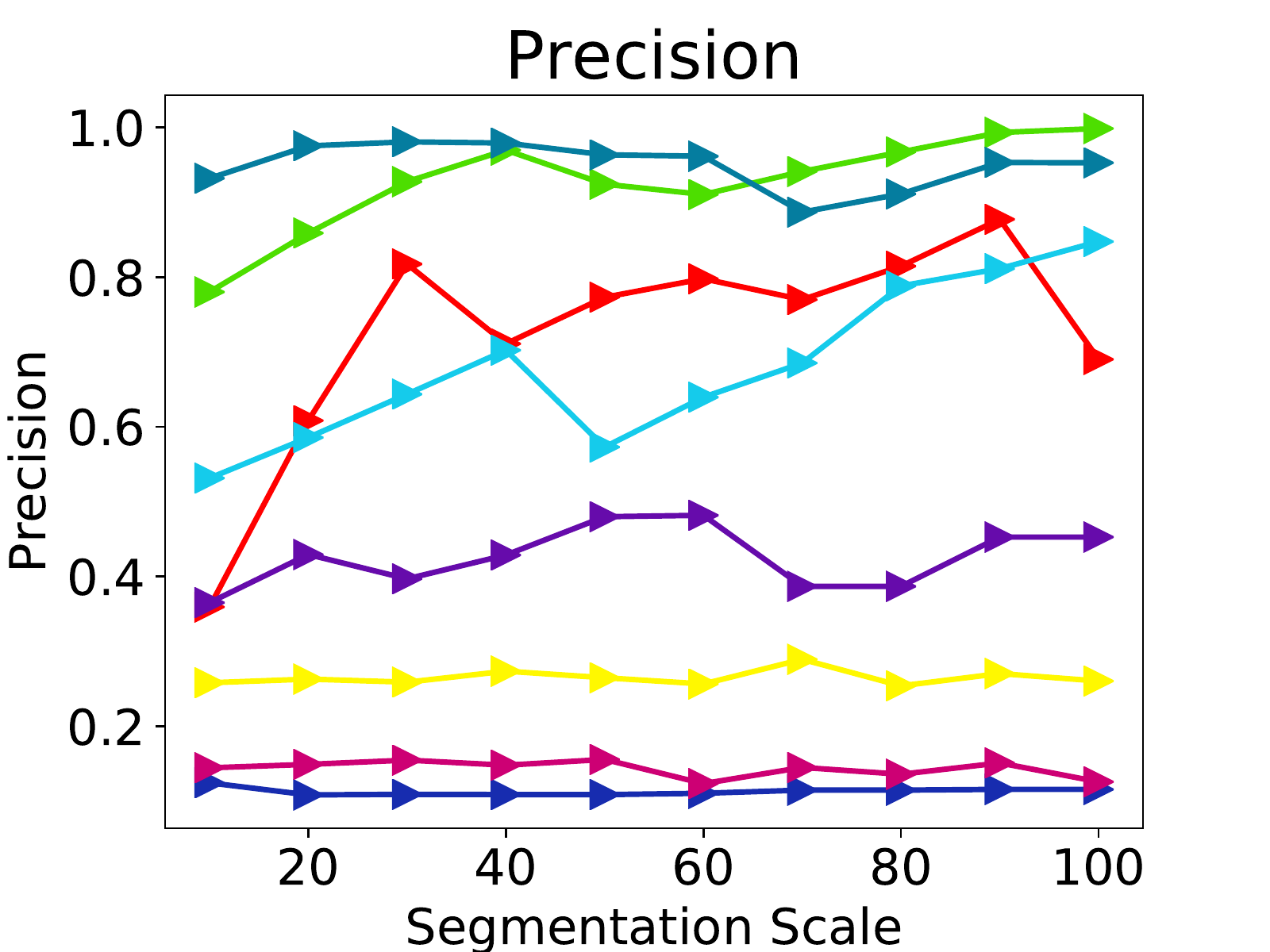}   
        \end{minipage}%
    }
    \subfloat[] 
    {
        \begin{minipage}[!t]{0.31\textwidth}
            \centering      
            \includegraphics[width=1\textwidth]{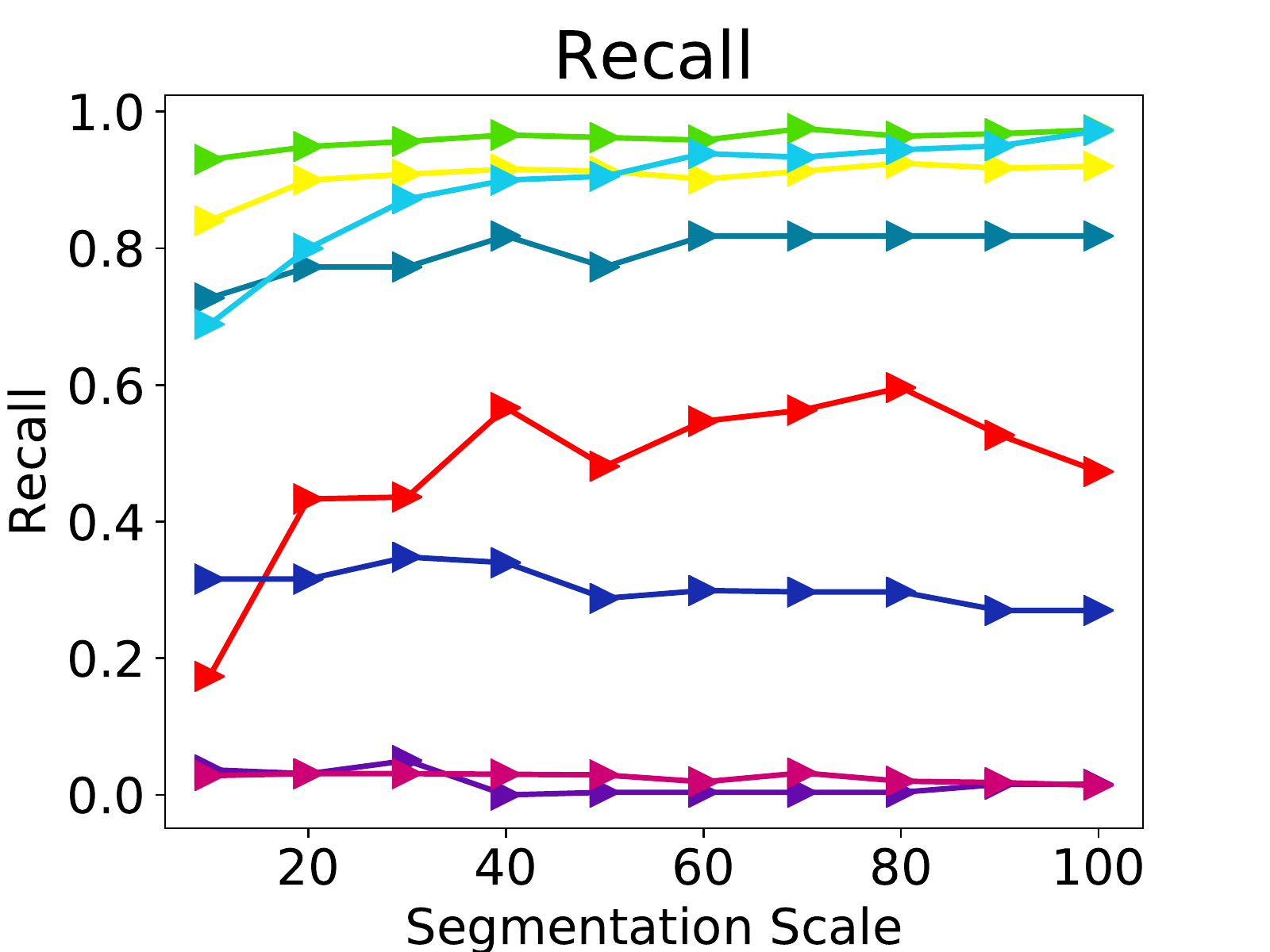}   
        \end{minipage}
    }
    \subfloat[] 
    {
        \begin{minipage}[!t]{0.37\textwidth}
            \centering      
            \includegraphics[width=1\textwidth]{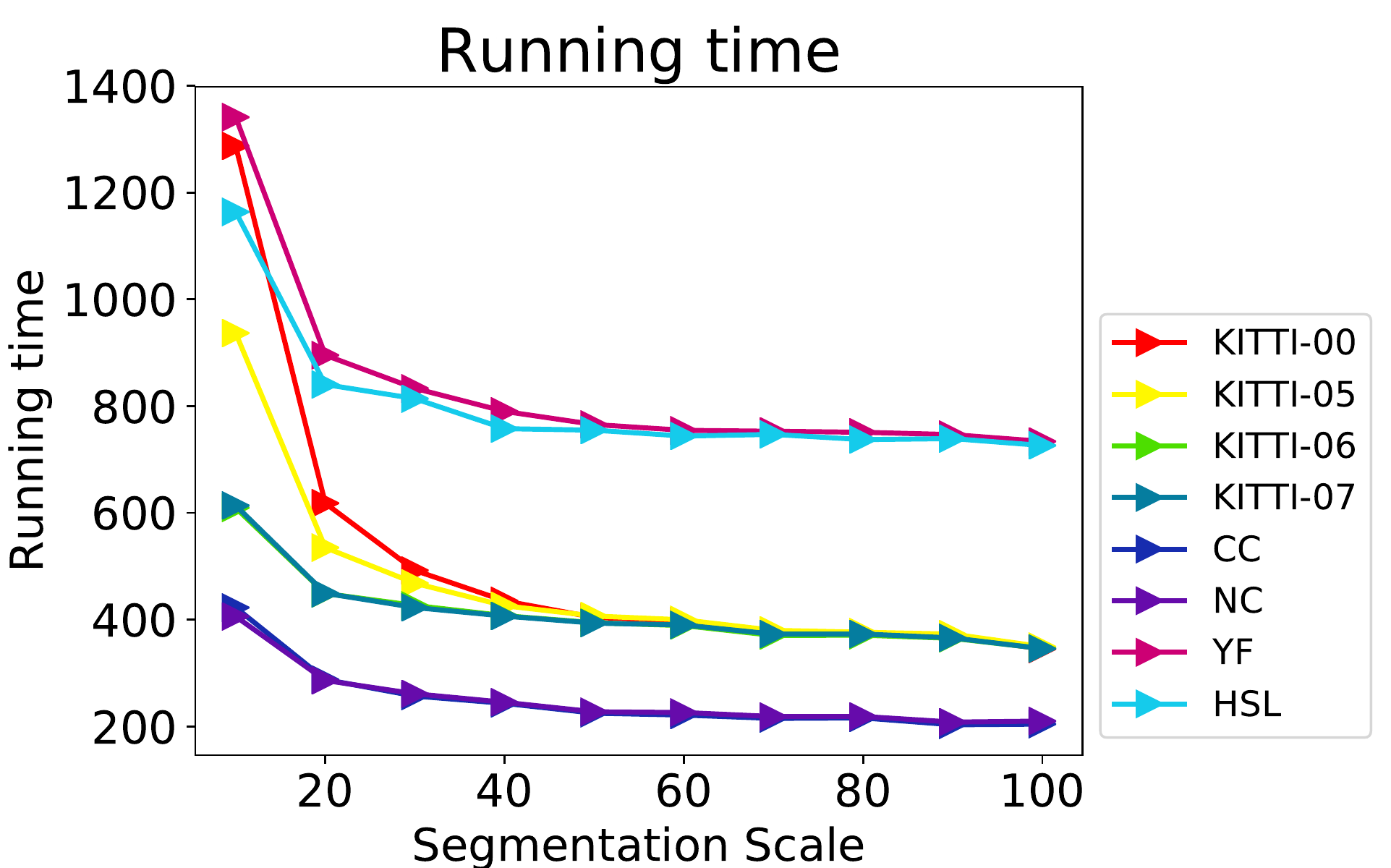}   
        \end{minipage}
    }    
    \caption{The performance of the LSGD with different segmentation scales in terms of maximum precision (\%), maximum recall (\%) and running time (ms) on the eight image sequences.} 
    \label{fig:exp1}  
\end{figure*}
In this section, we explored the impact of segmentation scale $S_p$ on the maximum precision rate and maximum recall rate at 100\% recall and 100\% precision, respectively, for the LSGDDN-LCD and conducted a series of experiments by gradually increasing $S_p$ from 10 to 100 with a step size of 10. As shown in Figure \ref{fig:exp1}(a) and (b), the precision and recall rates of all sequences increased with $S_p$ before stabilization and finally maintained a good performance in most datasets. This behavior could be attributed to the fact that larger $S_p$ allowed for the inclusion of more scene factors in the same SGs. This was a key factor in accurately calculating image similarity. 
Additionally, the average running time gradually decreased with increasing $S_p$ before stabilization, and the running time of sequences with uniform image size became consistent in the end. This was because that feature numbers decreased gradually with increasing $S_p$, and the feature numbers had a direct impact on the running time of the image similarity calculate process. Therefore, as the $S_p$ gradually increased, the major portion of the LCD execution time consumption shifted from similarity calculation to feature extraction. Ultimately, the running time depended on the original image size. 
Based on Figure \ref{fig:exp1}, we selected $S_p$ = 40 as the optimal segmentation scale in subsequent experiments. Using this value, the system keep the best achieving results.

\subsection{Effectiveness of the Dynamic Nodes}
\begin{table*}[!t]
\centering
 \caption{Performance of dynamic nodes approach on eight datasets} 
 \begin{tabular}{c c c c c c c c c c c} 
 \hline
 \multicolumn{1}{c}{\multirow{2}*{\textbf{Dateset}}}&\multicolumn{1}{c}{\multirow{2}*{\bm{$\beta$}}}&\multicolumn{1}{c}{\multirow{2}*{\textbf{\#Nodes}}}&\multicolumn{4}{c}{LSGD + dynamic nodes}&\multicolumn{4}{c}{LSGD}\\
 \cmidrule(r){4-7}
 \cmidrule(r){8-11}
 ~ & ~ & ~ & Times & R@1 & R@5 & R@10 & Times & R@1 & R@5 & R@10\\ [0.5ex] 
 \hline
 \hline
 \multirow{3}*{KITTI-00} & 0.9 & 91 & 363.267 & 66.27 & 86.68 & 88.46 & \multirow{3}*{435.176} & \multirow{3}*{98.81} & \multirow{3}*{99.26} & \multirow{3}*{99.55}\\
 ~ & 1.0 & 150 & \textcolor{red}{360.015} & 71.59 & \textcolor{blue}{94.82} & \textcolor{blue}{96.30} & ~ & ~ & ~ & ~\\
 ~ & 1.1 & 1972 & 364.504 & \textcolor{blue}{81.65} & 81.65 & 81.65 & ~ & ~ & ~ & ~\\
 \hline
 \multirow{3}*{KITTI-05} & 0.9 & 65 & 377.541 & 40.86 & 61.44 & \textcolor{blue}{66.37} & \multirow{3}*{426.042} & \multirow{3}*{96.52} & \multirow{3}*{97.39} & \multirow{3}*{97.39}\\
 ~ & 1.0 & 113 & \textcolor{red}{366.545} & 46.95 & 61.73 & 65.50 & ~ & ~ & ~ & ~\\
 ~ & 1.1 & 1246 & 380.418 & \textcolor{blue}{62.89} & \textcolor{blue}{62.89} & 62.89 & ~ & ~ & ~ & ~\\
 \hline
 \multirow{3}*{KITTI-06} & 0.9 & 22 & 359.468 & 70.15 & 91.08 & 92.63 & \multirow{3}*{407.396} & \multirow{3}*{100.00} & \multirow{3}*{100.00} & \multirow{3}*{100.00}\\
 ~ & 1.0 & 31 & \textcolor{red}{344.248} & 67.05 & \textcolor{blue}{92.24} & \textcolor{blue}{93.79} & ~ & ~ & ~ & ~\\
 ~ & 1.1 & 453 & 368.831 & \textcolor{blue}{70.93} & 71.31 & 71.31 & ~ & ~ & ~ & ~\\
 \hline
 \multirow{3}*{KITTI-07} & 0.9 & 28 & 410.403 & 0.00 & 9.09 & 9.09 & \multirow{3}*{406.689} & \multirow{3}*{72.72} & \multirow{3}*{72.72} & \multirow{3}*{81.81}\\
 ~ & 1.0 & 47 & \textcolor{red}{406.105} & 0.00 & \textcolor{blue}{18.18} & \textcolor{blue}{27.27} & ~ & ~ & ~ & ~\\
 ~ & 1.1 & 537 & 429.395 & \textcolor{blue}{9.09} & 9.09 & 9.09 & ~ & ~ & ~ & ~\\
 \hline
 \hline
 \multirow{3}*{City Center} & 0.9 & 7 & 299.273 & \textcolor{blue}{52.05} & \textcolor{blue}{86.00} & \textcolor{blue}{94.44} & \multirow{3}*{243.247} & \multirow{3}*{89.30} & \multirow{3}*{96.70} & \multirow{3}*{98.14}\\
 ~ & 1.0 & 11 & 257.258 & \textcolor{blue}{52.05} & 85.59 & 93.62 & ~ & ~ & ~ & ~\\
 ~ & 1.1 & 353 & \textcolor{red}{215.809} & 39.91 & 45.47 & 48.35 & ~ & ~ & ~ & ~\\
 \hline
 \multirow{3}*{New College} & 0.9 & 8 & 196.907 & 72.75 & \textcolor{blue}{97.43} & \textcolor{blue}{98.71} & \multirow{3}*{245.241} & \multirow{3}*{89.74} & \multirow{3}*{99.03} & \multirow{3}*{99.35}\\
 ~ & 1.0 & 9 & 211.941 & \textcolor{blue}{73.39} & \textcolor{blue}{97.43} & 98.07 & ~ & ~ & ~ & ~\\
 ~ & 1.1 & 238 & \textcolor{red}{187.406} & 61.53 & 76.60 & 77.24 & ~ & ~ & ~ & ~\\
 \hline
 \hline
 \multirow{3}*{YF} & 0.9 & 88 & 720.484 & 21.69 & \textcolor{blue}{67.33} & \textcolor{blue}{79.80} & \multirow{3}*{790.051} & \multirow{3}*{75.81} & \multirow{3}*{85.53} & \multirow{3}*{93.01}\\
 ~ & 1.0 & 122 & 708.232 & \textcolor{blue}{24.18} & 65.08 & 76.55 & ~ & ~ & ~ & ~\\
 ~ & 1.1 & 648 & \textcolor{red}{704.158} & 9.72 & 9.97 & 9.97 & ~ & ~ & ~ & ~\\
 \hline
 \multirow{3}*{HSL} & 0.9 & 62 & 724.432 & 34.52 & \textcolor{blue}{71.42} & \textcolor{blue}{75.00} & \multirow{3}*{757.712} & \multirow{3}*{92.85} & \multirow{3}*{92.85} & \multirow{3}*{94.04}\\
 ~ & 1.0 & 104 & 768.909 & \textcolor{blue}{39.28} & 64.28 & 64.28 & ~ & ~ & ~ & ~\\
 ~ & 1.1 & 459 & \textcolor{red}{692.633} & 23.80 & 25.00 & 25.00 & ~ & ~ & ~ & ~\\
 \hline
\end{tabular}
 \begin{tablenotes} 
        \item The red and blue colors represent the best results in the terms of running time and recall performance in method with $dynamic$ $nodes$, respectively.
    \end{tablenotes} 
\label{table:exp2}
\end{table*}

To evaluate the effectiveness of the $dynamic$ $nodes$, we conducted a comparative experiment for LCD methods that utilized various databases with severe illumination or view point changes and dynamic objects, i.e., City Center, New College \cite{cummins2008fab}, KITTI \cite{geiger2012we}, and NEU. We evaluated the recognition performance mainly based on Recall@N, whereby a query was regarded as correctly localized if at least one of the top N retrieved database images was within the ground truth tolerance. We recorded the running time and recognition performance of the LSGD using the $dynamic$ $nodes$ with various threshold values of $\beta$, as described in Section \ref{sec:33}, and the results were presented in Table \ref{table:exp2}. To analyze the influence of $\beta$ values on the $dynamic$ $nodes$ in terms of running time, we recorded node numbers generated with $\beta$ equal to 0.9, 1.0, and 1.1. We observed that the execution time per frame was the lowest when $\beta$ was set to 1.0 or 1.1. Increasing $\beta$ values resulted in fewer images per node, which affected both the similarity score calculation and the numbers of backtracking nodes, influencing the execution times for the system.
Therefore, selecting the optimal image numbers in each node, i.e., the optimal $\beta$, could lead to the most efficient LCD process.
Subsequently, we compared the performance of the LSGD with and without $dynamic$ $nodes$, as shown in Table \ref{table:exp2}. 
It was noteworthy that the LSGD using $dynamic$ $nodes$ achieved improvement in terms of average execution time across all datasets, i.e., over 11.72\%, 4.06\%, and 4.43\% for the KITTI, FAB-MAP, and NEU datasets, respectively, when compared to the method of LSGD without $dynamic$ $nodes$, with $\beta$ = 1.0.

Furthermore, we studied how $\beta$ value affected the recognition performance of the LCD by comparing the LSGD with the $dynamic$ $nodes$ with various $\beta$ values and a standard LSGD without $dynamic$ $nodes$. It is noteworthy that $dynamic$ $nodes$ only retained previously seen images for similarity calculation with the current query images, and the images that obtained after the current image were not included in the nodes. Therefore, the recognition of image functionality in this part was based solely on the previous images' recognition function of the current image.
As shown in Table \ref{table:exp2}, the best performance was generally achieved with $\beta$ equal to 0.9 or 1.0, due to the fact that the $\beta$ directly affected image numbers in each node as mentioned above. Larger $\beta$ values usually resulted in smaller image numbers in each node. This might cause the query image to be collected in nearby similar $dynamic$ $nodes$. For traditional LCD methods, there was a degradation in performance compared to the $dynamic$ $nodes$ since highly similar nodes attracted the query image before finding the correct pair image.


The above result showed that, the $dynamic$ $nodes$ method could significantly improve the running efficiency while maintaining high-level or similar recognition performance as in traditional LCD methods.

\subsection{Comparative Results on Various LCD Algorithms}
\label{equ:exp1}
\begin{figure*}[!t]  
    \centering    
    \subfloat[] 
    {
        \begin{minipage}[!t]{0.25\textwidth}
            \centering          
            \includegraphics[width=\textwidth]{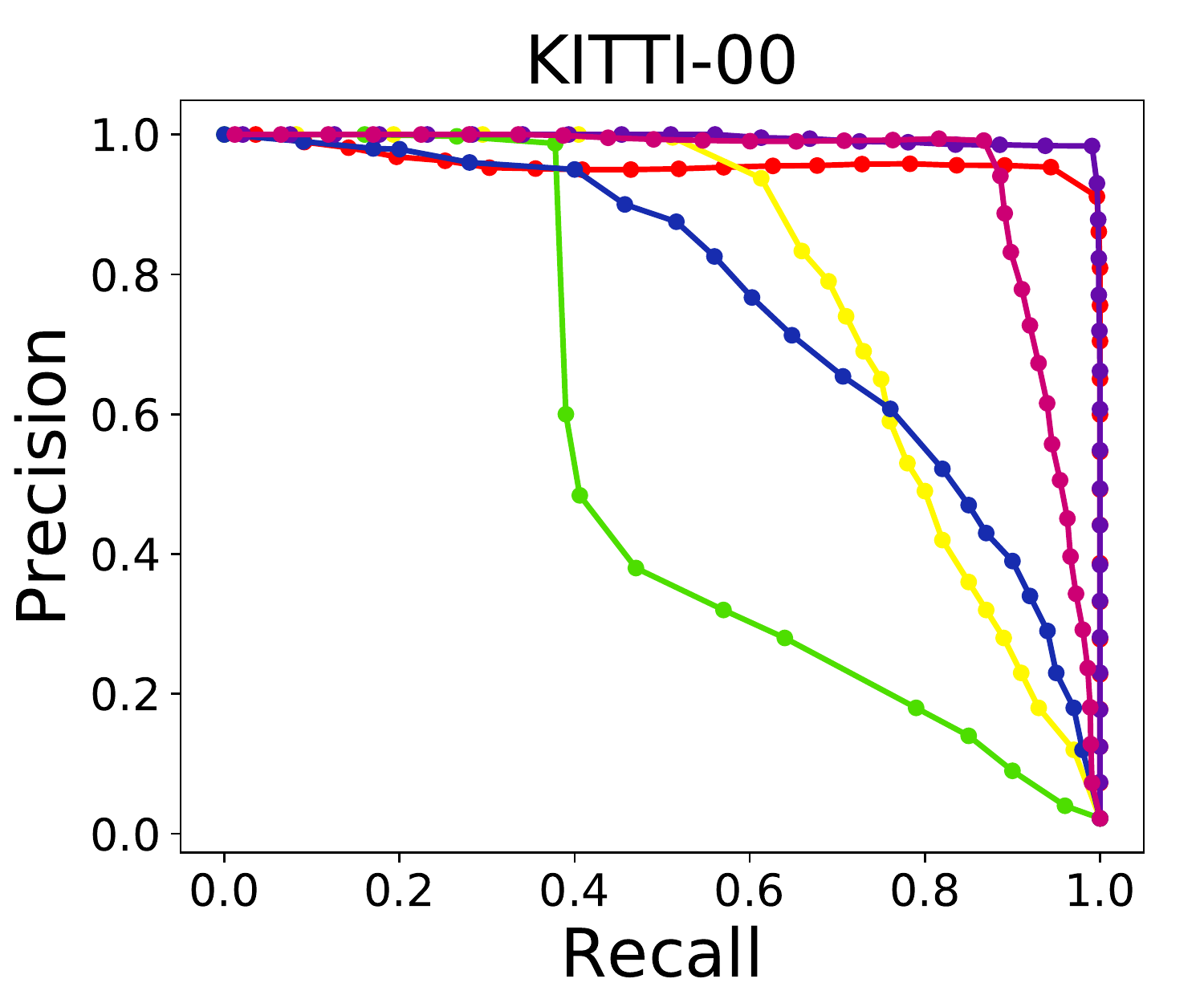}   
        \end{minipage}%
    }
    \subfloat[] 
    {
        \begin{minipage}[!t]{0.25\textwidth}
            \centering      
            \includegraphics[width=1\textwidth]{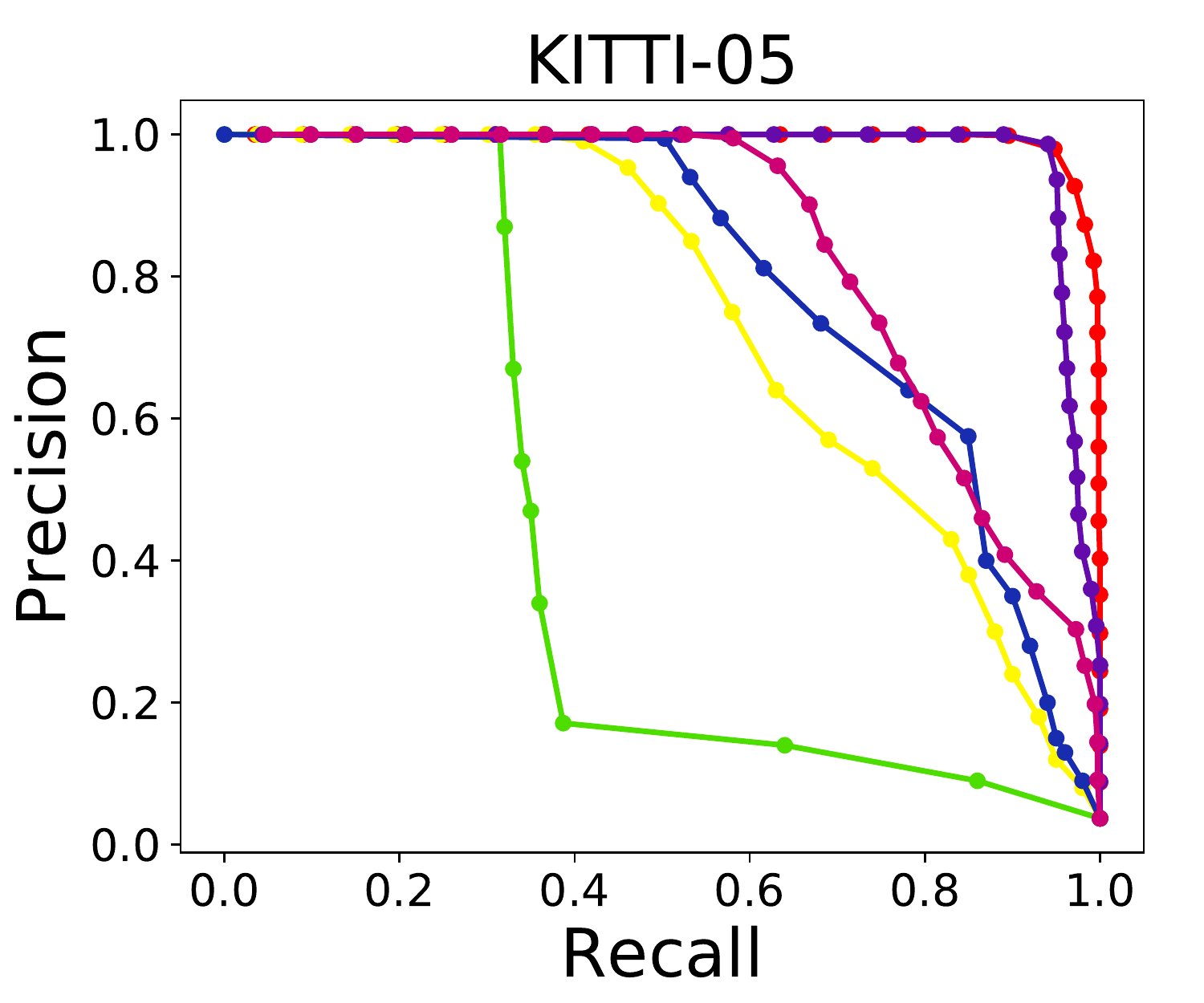}   
        \end{minipage}
    }
    \subfloat[] 
    {
        \begin{minipage}[!t]{0.25\textwidth}
            \centering      
            \includegraphics[width=1\textwidth]{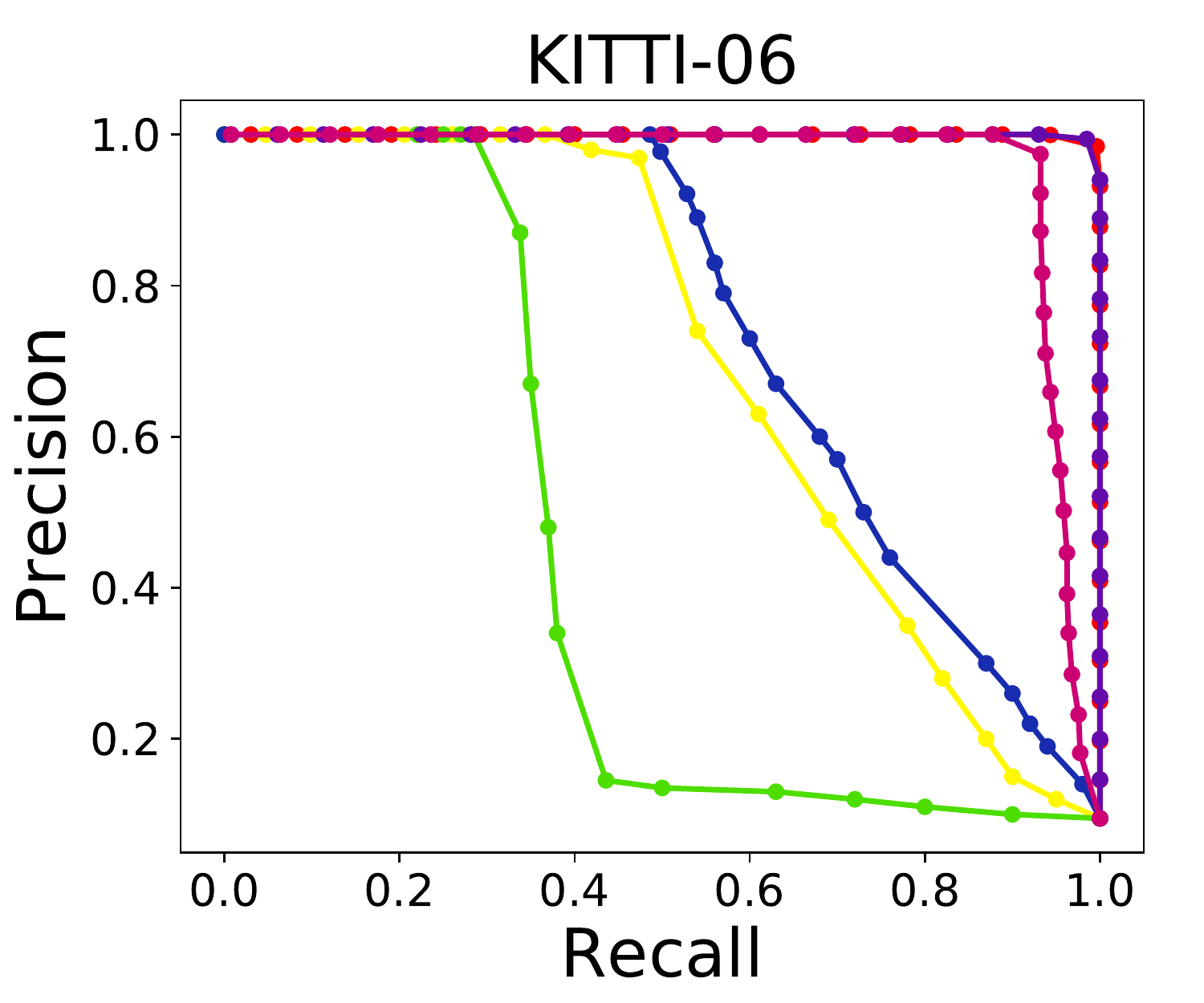}   
        \end{minipage}
    }    
    \subfloat[] 
    {
        \begin{minipage}[!t]{0.25\textwidth}
            \centering      
            \includegraphics[width=1\textwidth]{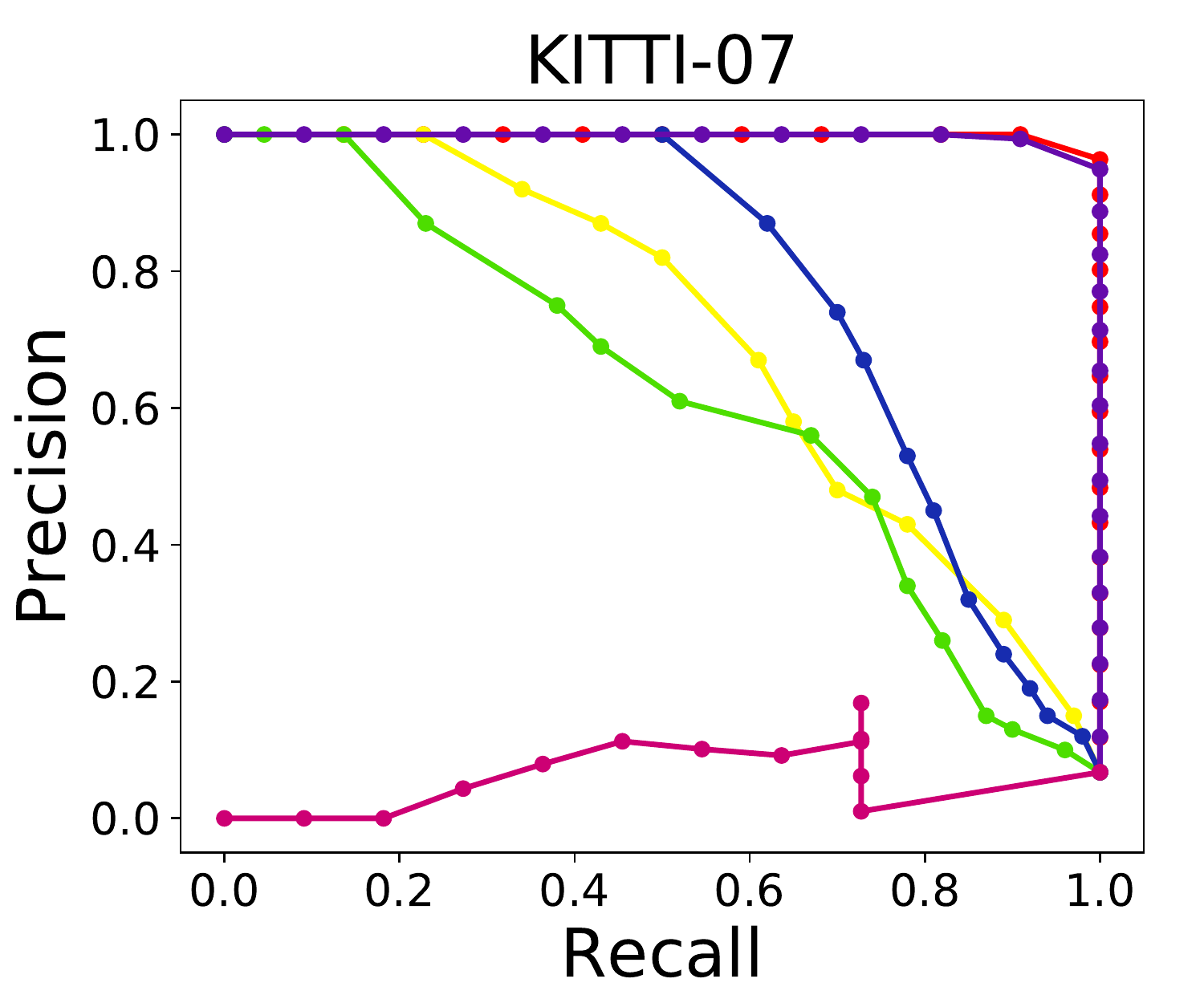}   
        \end{minipage}
    }

    \subfloat[] 
    {
        \begin{minipage}[!t]{0.25\textwidth}
            \centering          
            \includegraphics[width=\textwidth]{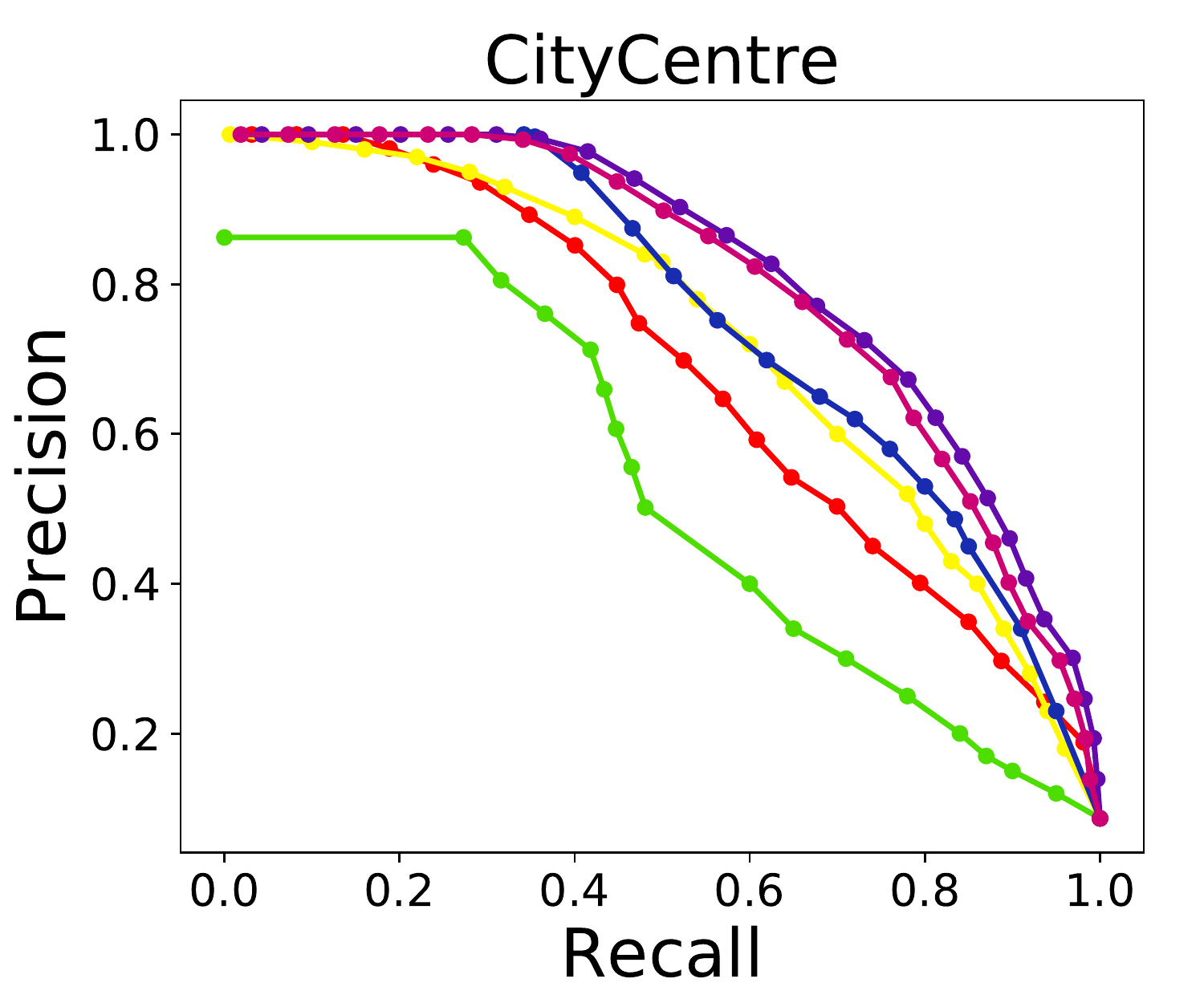}   
        \end{minipage}%
    }
    \subfloat[] 
    {
        \begin{minipage}[!t]{0.25\textwidth}
            \centering      
            \includegraphics[width=1\textwidth]{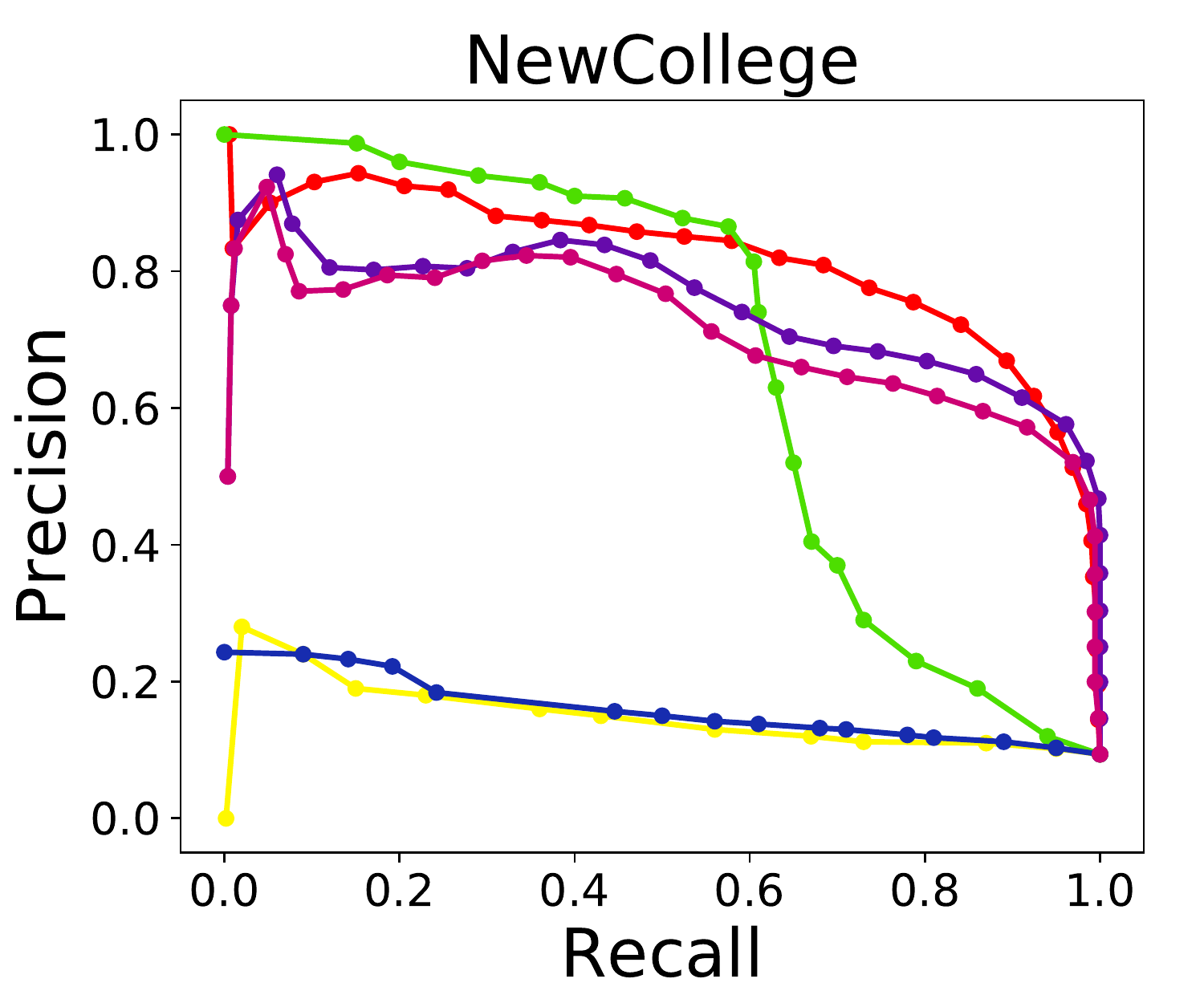}   
        \end{minipage}
    }
    \subfloat[] 
    {
        \begin{minipage}[!t]{0.25\textwidth}
            \centering      
            \includegraphics[width=1\textwidth]{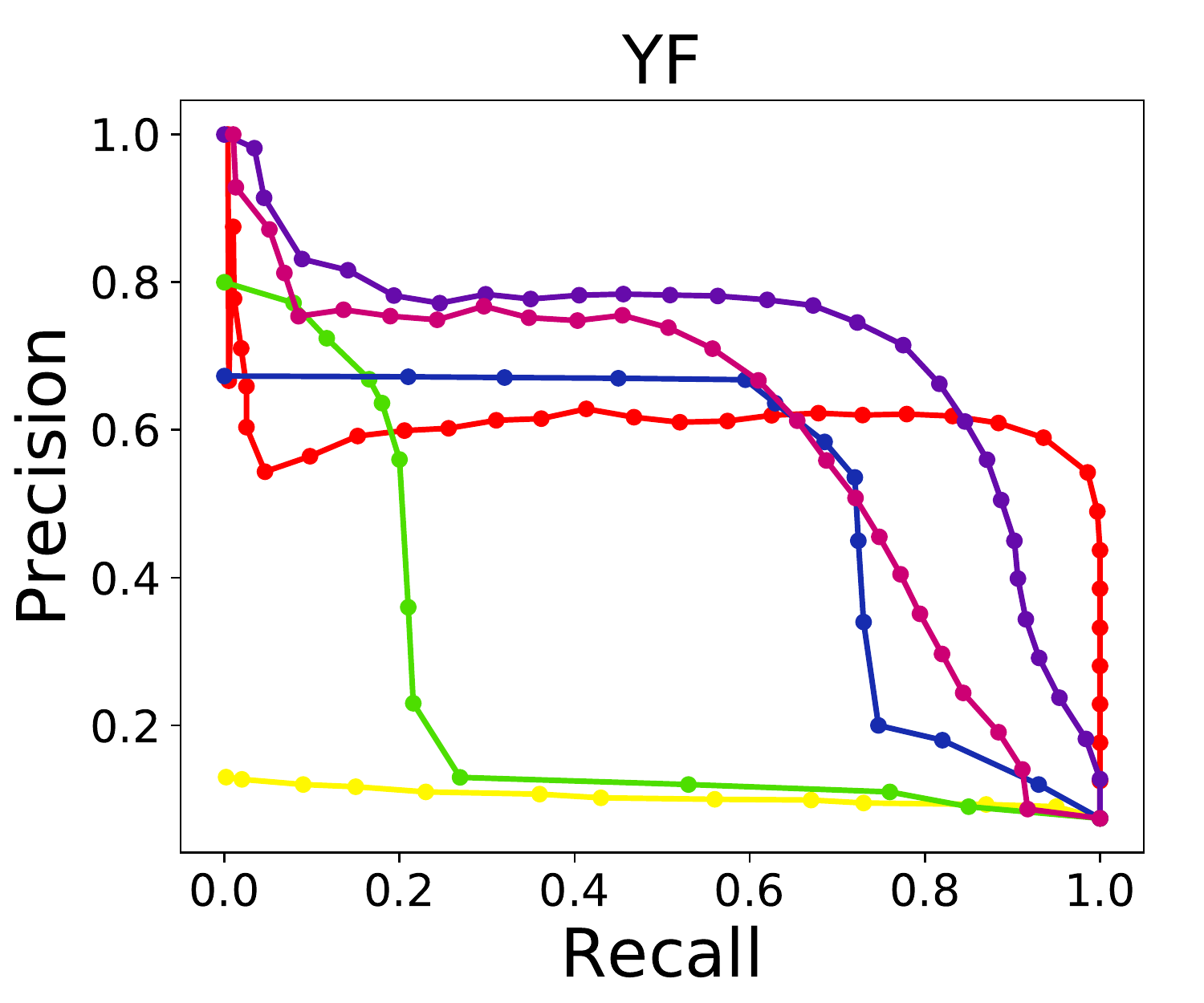}   
        \end{minipage}
    }    
    \subfloat[] 
    {
        \begin{minipage}[!t]{0.25\textwidth}
            \centering      
            \includegraphics[width=1\textwidth]{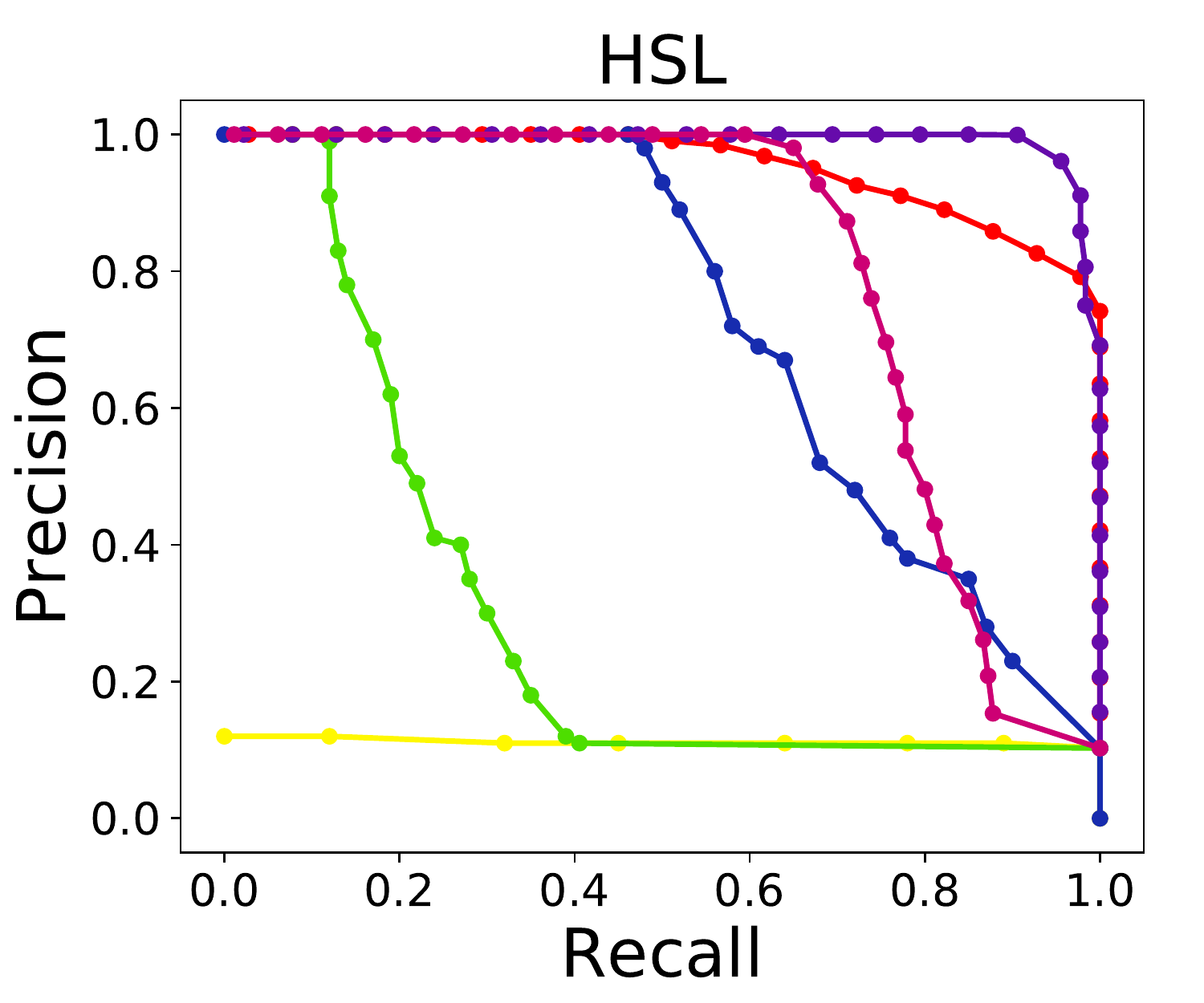}   
        \end{minipage}
    }
    
    {
        \begin{minipage}[!t]{0.7\textwidth}
            \centering      
            \includegraphics[width=1\textwidth]{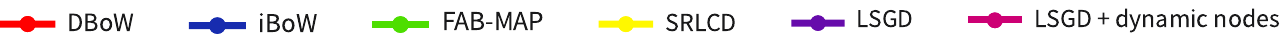}   
        \end{minipage}
    }
    \caption{The comparison of the precision-recall curves between LSGD and four SOTA LCD methods, i.e., DBoW, iBoW, FAB-MAP and SRLCD on the outdoor and indoor sequences. } 
    \label{fig:exp2}  
\end{figure*}
In this section, we presented a comprehensive evaluation of the LSGDDN-LCD method by providing both quantitative and qualitative analysis of LCD precision-recall and overall performance compared to four SOTA LCD algorithms.

We first compared them on eight datasets using precision-recall curves, as shown in Figure \ref{fig:exp2}. The results indicated that the LSGD outperformed other LCD methods on these datasets, especially for outdoor KITTI and FAB-MAP datasets. This was because that the key scene factors used for recognition in outdoor scenes could be effectively divided by grid segmentation based models. 

Figure \ref{fig:exp2} (d) showed performance degradation of the LSGD method with $dynamic$ $nodes$ on KITTI-07, mainly due to the scattered distribution of correct loop closure images in the dataset, making it hard to include correct loop closure images in the same nodes. In contrast, correct loop closure images were tightly distributed in KITTI-06, City Centre, and New College with $dynamic$ $nodes$ adjacent to each other, wherein the performance of the LSGD method with $dynamic$ $nodes$ framework almost reached the level of standard LSGD, as shown in Figure \ref{fig:exp2} (c), (e), and (f).

\begin{figure*}[!t]  
    \centering    
    \centering          
    \includegraphics[width=\textwidth]{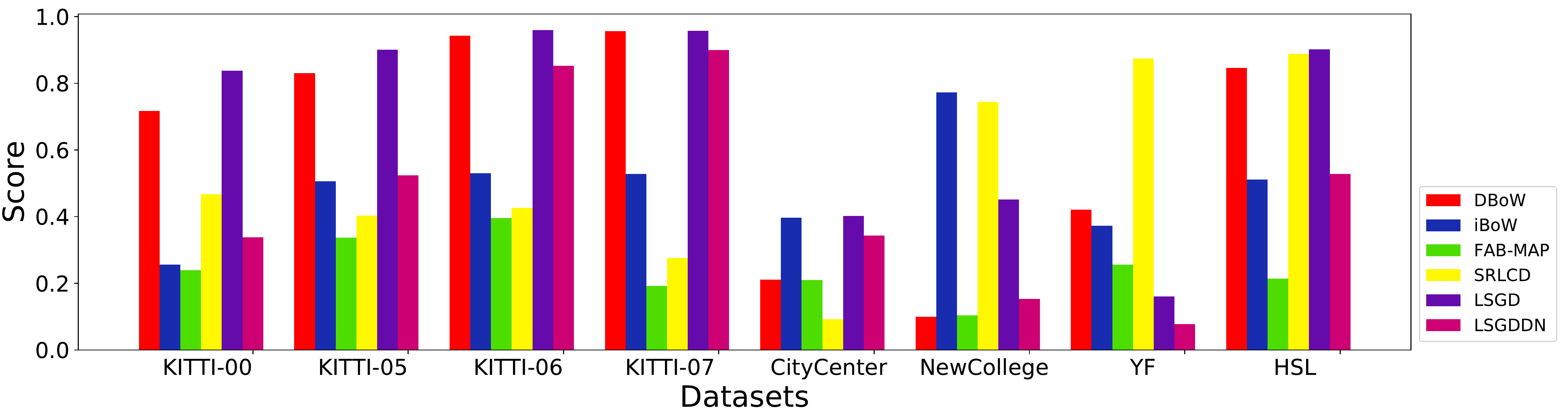}   
    \caption{The $PRT$ evaluation for all LCD approaches on eight datasets. ($\omega$ = 10.0)} 
    \label{fig:exp3}  
\end{figure*}
To provide a more quantitative analysis of the performance of the LSGDDN-LCD, we further evaluated the system's precision, recall, and running time with a comprehensive evaluation score $PRT$, defined as
\begin{equation*}
    PRT = AUC / (1 + \omega * Running\, Time(s))
\end{equation*}

Here $AUC$ denoted area under the precision-recall curves, $\omega$ was set to 10.0 in this work. Figure \ref{fig:exp3} indicated that the LSGD method outperformed all other LCD methods in the KITTI dataset. Particularly, the LSGD achieved an average improvement of 146.1\% and 94.1\% in $PRT$, compared with SOTA LCD approaches in the KITTI-00 and KITTI-05 sequences.
This was caused by two facts. Firstly, the LSGD method produced the best precision-recall performance in outdoor KITTI datasets as shown in Figure \ref{fig:exp2}. Secondly, the LSGD maintained relatively stable running time as the image numbers increased, thereby providing better average execution time performance in the KITTI dataset with a large image numbers.
As for the LSGD method with $dynamic$ $nodes$, the comprehensive performance only achieved average levels among all other methods in most cases. This attributed to the fact that optimizing running time might also reduce the algorithm's performance in the precision-recall trade-off for specific experimental data. Therefore, implementation methods should be carefully chosen according to specific tasks.

\section{Conclusions}
\label{sec:5}
In this letter, we presented a novel appearance-based loop closure detection approach, LSGDDN-LCD, that utilized patch-level features extracted based-on superior grids and local region pixel intensities to represent images, and a $dynamic$ $nodes$ framework was also introduced to organize previously visited images and selected candidate frames without the need to search through all images in the database. Qualitative and quantitative testings on four datasets with severe scene changes demonstrated the superiority of the LSGDDN-LCD over other SOTA appearance-based LCD approaches in precision-recall evaluation.
Comprehensive results demonstrated that the LSGD approach could effectively handle visual occlusion and environmental changes, enhancing the precision-recall performance and robustness of the LCD methods. 
By adopting the $dynamic$ $nodes$ strategy and selecting appropriate parameters, processing time could be significantly reduced with only a negligible impact on precision and recall.

In the future work, we plan to incorporate raw depth information from sensor inputs into the LSGDDN-LCD to enhance its understanding of the scenes and improve the performance of loop-closure recognition.

\bibliographystyle{ieeetr}
\bibliography{references}

\end{document}